\documentclass[letterpaper]{article} 
\usepackage{aaai2027}  
\nocopyright
\usepackage{amsmath}

\newcommand{\method}{ProTAGAD}

\usepackage[hyphens]{url}  
\usepackage{graphicx} 
\usepackage{natbib}  
\usepackage{caption} 
\usepackage{algorithm}
\usepackage{algorithmic}
\usepackage{amssymb}
\usepackage[dvipsnames, svgnames, x11names]{xcolor}
\usepackage{newfloat}
\usepackage{listings}
\DeclareCaptionStyle{ruled}{labelfont=normalfont,labelsep=colon,strut=off} 
\floatstyle{ruled}
\newfloat{listing}{tb}{lst}{}
\floatname{listing}{Listing}

\usepackage{booktabs}

\title{ProTAGAD: A Foundation Model for TAG Anomaly Detection with Decoupled Topological and Textual Prototypes}
\author{
    Ziyan Wang$^\dagger$, 
    Liwen Wu$^\dagger$, 
    Cheng Xie$^*$, 
    Song Gao, 
    Zhenli He, 
    Xin Jin
}

\affiliations{
    School of Software and AI, Yunnan University, Kunming, China\\
    wangziyan\_iaab@stu.ynu.edu.cn, lwwu@ynu.edu.cn, xiecheng@ynu.edu.cn,\\ gaos@ynu.edu.cn, hezl@ynu.edu.cn, xinjin@ynu.edu.cn
}

\begin{document}

\maketitle

\begin{abstract}
Text-Attributed Graphs (TAGs), endowed with abundant textual content along with topological structures, have emerged as a versatile backbone for real-world anomaly detection spanning large language model security, social network moderation, and cyber threat identification. 
Unlike conventional Graph Anomaly Detection (GAD), which relies primarily on structural irregularities, TAG anomaly detection must jointly leverage both topological patterns and fine-grained textual semantics to capture nuanced anomalous behaviors. 
The current GNN-based anomaly detectors adopt holistic message-passing schemes that indiscriminately fuse structural proximity and textual semantics during propagation, leading to deep cross-modality coupling. 
This entanglement acts as a noise amplifier, obscuring subtle anomalous signals and directly giving rise to the \textbf{\textit{Blurred-Anomaly-Boundary (BAB)}} issue by rendering normal-anomalous decision boundaries poorly separable.
This challenge is further amplified for graph foundation models that require robust cross-domain generalization. 
To bridge this gap, we introduce a novel foundation model for TAG anomaly detection featuring decoupled topological and textual prototypes. 
Our framework constructs dual prototype banks to independently model structural normality and semantic consistency, effectively isolating anomaly cues that are otherwise diluted during coupled aggregation.
Extensive experiments across 14 diverse benchmark datasets demonstrate that our method consistently achieves state-of-the-art performance in cross-domain settings. 
Notably, the ablation studies further corroborate the prevalence of the \textbf{\textit{BAB}}  issue in conventional coupled TAG anomaly detectors, and show that our decoupled prototype design effectively mitigates this challenge.
\end{abstract}


\section{Introduction}
\begin{figure}[t]
    \centering
    \includegraphics[width=\linewidth]{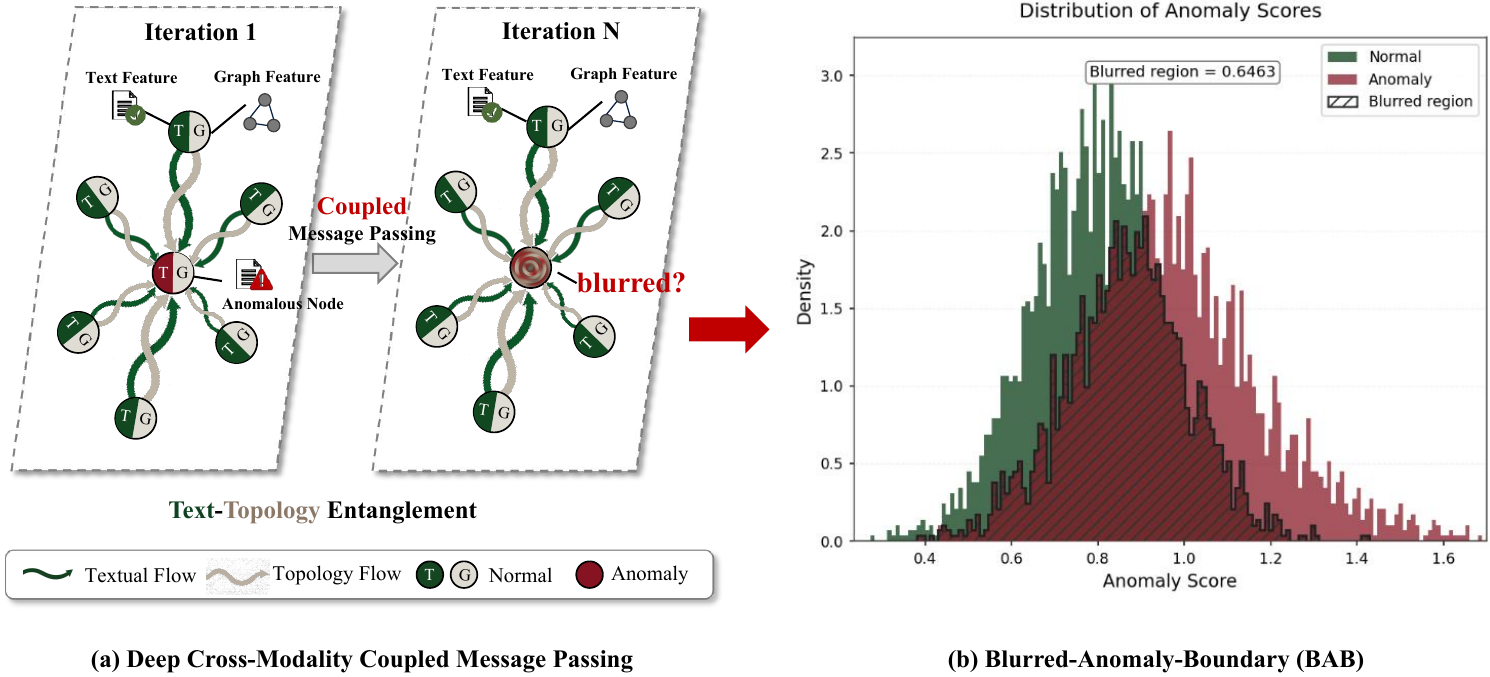}
    \caption{Motivation: deep cross-modality coupling obscures subtle anomaly cues, leading to the Blurred-Anomaly-Boundary (BAB) problem.}
    \label{fig:motivation}
\end{figure}
Graph anomaly detection (GAD) aims to identify nodes that deviate from dominant graph patterns and has been widely applied to fraud detection, social network moderation, cybersecurity, and recommendation systems~\cite{ding2019deep,liu2021anomaly,ma2021comprehensive,qiao2025deep}. In real-world graphs, nodes are often associated with rich textual content in addition to relational structures, such as paper abstracts, product descriptions, and user posts. Such data are commonly formulated as text-attributed graphs (TAGs)~\cite{zhu2026graph,bandyopadhyay2019outlier}, where anomalies may arise from irregular connections, subtle semantic inconsistencies, or both. Effective TAG anomaly detection therefore requires jointly exploiting textual semantics and graph topology~\cite{yang2021graphformers,zhao2022learning}.

Existing GAD methods are primarily designed for numerical attributes and structural irregularities. Although recent TAGAD methods employ language models to encode raw text, most still rely on holistic GNN message passing to jointly propagate textual and topological information. As illustrated in Figure~\ref{fig:motivation}, this coupled propagation progressively smooths or distorts subtle anomaly cues, reducing the distinction between normal and anomalous nodes and thereby giving rise to the \textit{Blurred-Anomaly-Boundary} (\textit{BAB}) issue.

Recent graph pre-training and prompt-learning methods have explored transferable graph representations across tasks and domains~\cite{hou2022graphmae,sun2022gppt,liu2023graphprompt,liu2025graph,xiong2025noisehgnn}. However, the \textit{BAB} issue becomes more severe in the generalist setting, where one model is trained on multiple source graphs and directly applied to unseen domains. Cross-domain variations in textual semantics and connectivity patterns make jointly propagated representations prone to retaining domain-specific neighborhood information, further weakening anomaly cues under distribution shifts and hindering zero-shot generalization.

To address this issue, we propose \textbf{ProTAGAD}, a prototype-based foundation model that decouples textual and topological representation learning. ProTAGAD separately learns transferable textual anomaly prototypes and topological normality prototypes without exchanging hidden representations, and combines their anomaly scores only at the decision level. This design avoids cross-modal interference while preserving the complementary anomaly evidence of both modalities.

We evaluate ProTAGAD on 14 TAG datasets under a source-to-target zero-shot protocol. ProTAGAD achieves the best AUROC on seven of eight unseen target graphs, with an average rank of 1.12, demonstrating strong cross-domain generalization. Additional analyses further reveal the complementary roles of textual and topological prototypes and verify that decoupled prototype modeling effectively alleviates the \textit{BAB} issue.

Our main contributions are summarized as follows:
\begin{itemize}
\item We formally identify the Blurred-Anomaly-Boundary (\textit{BAB}) issue in generalist TAG anomaly detection, attributing its root cause to entangled cross-modal fusion via holistic GNN message passing.
\item We propose ProTAGAD, a foundation model for TAG anomaly detection that leverages decoupled textual anomaly prototypes and topological normality prototypes to isolate modality-specific anomaly evidence without cross-modal interference.
\item Extensive experiments on 14 diverse TAG benchmarks achieve state-of-the-art zero-shot cross-domain performance; ablations further verify dual prototype complementarity and empirically confirm both the prevalence of the BAB issue and the efficacy of our decoupled design.
\end{itemize}

\section{Related Work}

\subsection{Graph Anomaly Detection}
GAD aims to identify nodes that deviate from the dominant attribute or structural patterns~\cite{wang2023cross}. Existing methods mainly include reconstruction-, self-supervised-, spectral-, affinity-, and augmentation-based approaches. Reconstruction-based methods detect anomalies through the reconstruction of attributes or structures~\cite{ding2019deep,fan2020anomalydae,luo2022comga,duan2026designated}. Self-supervised methods learn anomaly-sensitive node-context or neighborhood patterns, represented by CoLA~\cite{liu2021anomaly} and HCM-A~\cite{huang2022hop}. Spectral methods exploit high-frequency graph signals, including BWGNN~\cite{tang2022rethinking} and GHRN~\cite{gao2023addressing}, while affinity-based methods model one-class normality and suppress suspicious connections, such as TAM~\cite{qiao2023truncated} and GCTAM~\cite{ijcai2025p405}. Augmentation-based methods, including Semi-GGAD~\cite{qiao2024generative} and CAGAD~\cite{xiao2024counterfactual}, synthesize anomalies to enhance model robustness. However, these methods generally follow a one-for-one paradigm with fixed numerical attributes, requiring retraining on new graphs and overlooking the fine-grained semantics of raw text.

\subsection{Generalist Graph Anomaly Detection}

Generalist graph anomaly detection (GGAD) extends conventional GAD from graph-specific learning toward cross-domain generalization, aiming to train a unified detector that can be transferred to unseen graphs~\cite{pan2025survey,wang2023cross}. ARC~\cite{liu2024arc} introduces in-context learning for transferable anomaly detection, while UNPrompt~\cite{niu2024zero} and AnomalyGFM~\cite{qiao2025anomalygfm} explore zero/few-shot detection through unified prompts and graph foundation models. Recent studies further investigate domain shifts through invariant representation learning and prototype-based knowledge transfer, such as IA-GGAD~\cite{zhang2026ia}, DR-GGAD~\cite{fu2026drggad}, OWLEYE~\cite{zheng2026owleye}, and ProMoS~\cite{xu2026generalist}.

However, existing GGAD methods generally focus on numerical node representations and structural distribution shifts. The semantic information contained in raw texts is not explicitly modeled, and the interaction between textual and topological anomaly evidence remains unexplored, limiting their applicability to text-attributed graphs.

\subsection{Text-Attributed Graph Anomaly Detection}
Text-attributed Graph Anomaly Detection (TAGAD) identifies anomalous nodes by jointly modeling textual attributes and graph topology. Conventional methods typically encode texts into fixed-dimensional representations, which may miss subtle anomaly cues. Recent approaches improve textual modeling through contrastive learning or large language model reasoning. CMUCL~\cite{xu2025text} captures textual and structural inconsistencies through multi-scale contrastive learning, CoLL~\cite{xu2025court} extracts textual anomaly evidence through collaborative LLM reasoning, and TAG-AD~\cite{xu2025llm} studies realistic anomaly construction and retrieval-augmented zero-shot detection.

However, existing methods often entangle textual and topological information, causing semantic anomalies to be smoothed by neighborhood aggregation, while LLM-based approaches rarely model transferable structural normality. Most are also tailored to individual graphs. Instead, ProTAGAD learns decoupled textual and topological prototype banks and combines their scores for cross-domain detection on unseen graphs.

\begin{figure*}[t]
    \centering
    \includegraphics[width=0.95\textwidth]{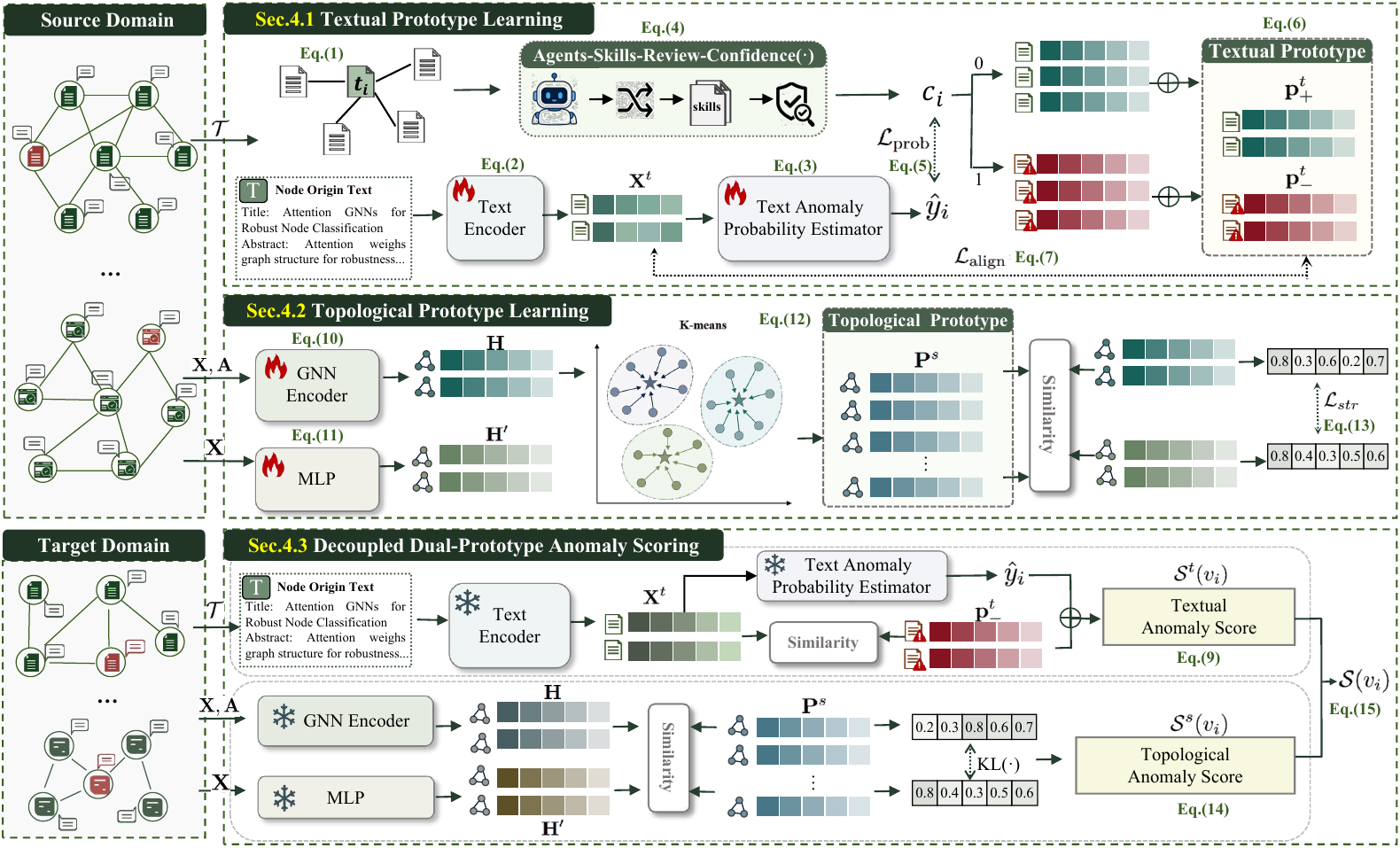}
    \caption{Overview of ProTAGAD.}
    \label{fig:framework}
\end{figure*}

\begin{figure*}[t]
    \centering
    \includegraphics[width=0.9\textwidth]{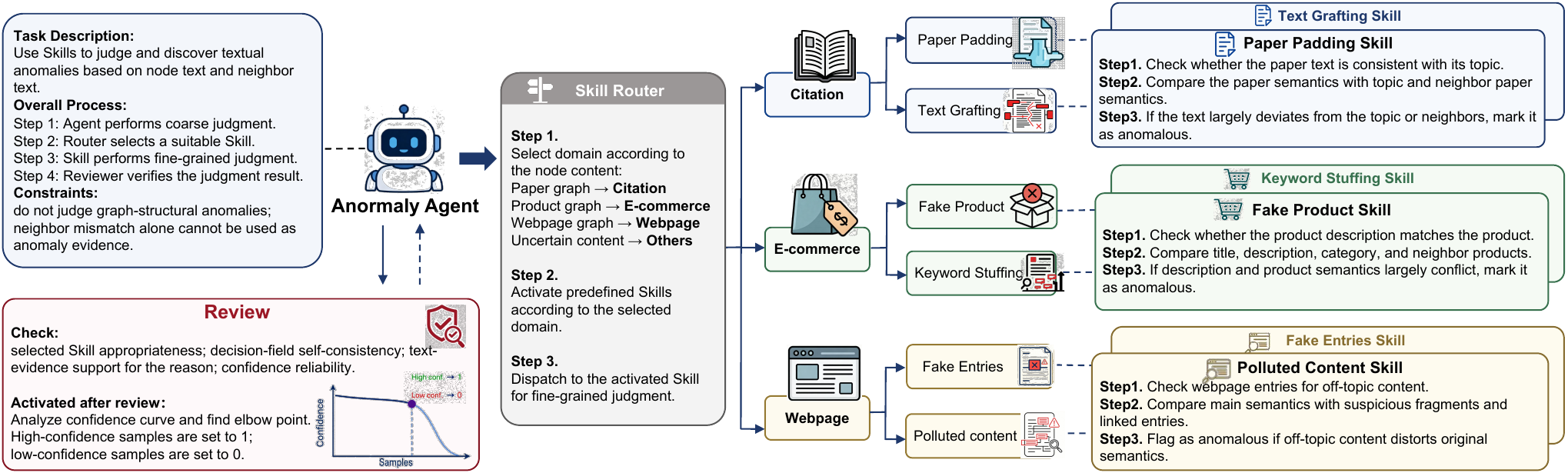}
    \caption{The overview of the function Agents-Skills-Review-Confidence($\cdot$). The input is textual subgraphs for each node. The output is the activated textual anomaly confidence $c_i$ for each node.}
    \label{fig:agent-skill}
\end{figure*}

\section{Preliminaries}
\paragraph{Notations. }
Let
$\mathcal{G}=(\mathcal{V},\mathbf{A},\mathcal{T},\mathbf{X})$
denote a text-attributed graph, where
$\mathcal{V}=\{v_1,\ldots,v_N\}$ is the node set,
$\mathbf{A}\in\{0,1\}^{N\times N}$ is the adjacency matrix, and
$\mathcal{T}=\{t_i\}_{i=1}^{N}$ contains the raw textual attributes
associated with the nodes. An entry $\mathbf{A}_{ij}=1$ indicates an
edge between $v_i$ and $v_j$. The matrix $\mathbf{X}\in\mathbb{R}^{N\times d}$ denotes the initial node feature matrix, whose $i$-th row $\mathbf{x}_i\in\mathbb{R}^{d}$ is the $d$-dimensional feature vector associated with node $v_i$.

For each node $v_i$, we denote its one-hop neighborhood by
$\mathcal{N}_i$ and construct a local textual subgraph containing the
text of the target node and its neighbors:
\begin{equation}
\mathcal{G}^t_i
=
\{t_i\}
\cup
\{t_j\mid j\in\mathcal{N}_i\}.
\label{eq:text_subgraph}
\end{equation}

The local textual subgraph enables the agent to assess both the semantic content of the target node and its consistency with neighboring texts.

\paragraph{Generalist TAG Anomaly Detection.}
Given a collection of source-domain TAGs
$\mathcal{T}_{\mathrm{train}}
=\{\mathcal{G}^{(1)}_{\mathrm{train}},\ldots,
\mathcal{G}^{(n_s)}_{\mathrm{train}}\}$
and a disjoint collection of unseen target-domain TAGs
$\mathcal{T}_{\mathrm{test}}
=\{\mathcal{G}^{(1)}_{\mathrm{test}},\ldots,
\mathcal{G}^{(n_t)}_{\mathrm{test}}\}$,
where $n_s$ and $n_t$ denote the numbers of source and target graphs,
respectively, our goal is to learn a unified anomaly detector from
$\mathcal{T}_{\mathrm{train}}$ and directly generalize it to
$\mathcal{T}_{\mathrm{test}}$. During inference, the model parameters
are frozen, and no target-domain labels or additional fine-tuning are
available. For each node $v_i$ in a target graph, the detector outputs an anomaly score $\mathcal{S}(v_i)\in\mathbb{R}$, where a larger value indicates a higher likelihood of being anomalous.

\section{Methodology}
To mitigate the \textit{BAB} problem, we propose ProTAGAD, which decouples textual and topological representation learning by avoiding shared cross-modal message propagation. As illustrated in Figure~\ref{fig:framework}, ProTAGAD separately learns textual and topological prototypes and fuses their anomaly scores for zero-shot inference on unseen graphs.

\subsection{Textual Prototype Learning}
The textual module aims to extract transferable semantic anomaly
patterns from raw node texts. We first employ a text encoder
$f_{\mathrm{txt}}^{\Theta}$ to map the textual attributes into a
continuous representation space:
\begin{equation}
\mathbf{X}^t
=f_{txt}^{\Theta}\left(\mathcal{T}\right)
=
\left[
\begin{array}{c}
\mathbf{x}^t_1,\dots,\mathbf{x}^t_N
\end{array}
\right]^\top
\in\mathbb R^{N\times d_t},
\label{eq:text_encoder}
\end{equation}
where $\mathbf{x}^t_i$ denotes the textual representation of node $v_i$.

Based on these textual representations, we employ a lightweight text anomaly probability estimator to estimate the textual anomaly probability of each node:
\begin{equation}
    \hat{\mathbf{Y}} = f_{prob}^{{\Theta}}(\mathbf{X}^t) = [\hat{y}_1, \dots, \hat{y}_N]^\top,
\label{eq:Text_Anomaly_Probability}
\end{equation}
where $\hat{y}_i\in[0,1]$ denotes the estimated textual anomaly probability of node $v_i$.

Existing text encoders often emphasize general semantics, making fine-grained anomaly cues difficult to capture. To address this issue, we introduce the $\mathtt{Agents}\text{-}\mathtt{Skills}\text{-}\mathtt{Review}\text{-}\mathtt{Confidence}$ mechanism illustrated in Figure~\ref{fig:agent-skill}. Given a local textual subgraph $\mathcal{G}_i^t$, the anomaly agent first performs a coarse assessment. The Skill Router then identifies the graph domain, activates the corresponding domain-specific skill, and conducts fine-grained analysis of whether the target text is semantically consistent with its topic or content~\cite{schick2023toolformer}. The reviewer further verifies the decision consistency, supporting textual evidence, and confidence reliability~\cite{madaan2023self,gou2023critic}. After reviewing all nodes, we identify an elbow point in the anomaly confidence distribution as the threshold. Nodes with confidence scores above the threshold are assigned a pseudo-label of $1$, while the remaining nodes are assigned $0$. The resulting binary indicator for node $v_i$ is defined as
\begin{equation}
\begin{aligned}
c_i
=
\mathtt{Agents}\text{-}\mathtt{Skills}\text{-}\mathtt{Review}\text{-}\mathtt{Confidence}
\left(
\mathcal{G}^t_i
\right),
\label{eq:agent-skill-review-confidence}
\end{aligned}
\end{equation}
where $\mathbf{C}=[c_1,\ldots,c_N]^\top$, with $c_i=1$ indicating a high-confidence textual anomaly and $c_i=0$ indicating low textual anomaly confidence. These agent-derived binary indicators are used only as pseudo-labels to supervise the text anomaly probability estimator through the following binary cross-entropy loss:
\begin{equation}
\mathcal{L}_{\mathrm{prob}}
=
-\frac{1}{N}
\sum_{i=1}^{N}
\left[
c_i\log(\hat{y}_i)
+
(1-c_i)\log(1-\hat{y}_i)
\right].
\label{eq:loss_prob}
\end{equation}

We further use these pseudo-labels to partition the node representations into anomalous and normal groups, which are then aggregated to construct the corresponding textual prototypes:
\begin{equation}
\begin{aligned}
{\mathbf{p}^{t}_{-}}
\!=\!
\frac{1}{\|C\|_1}
\sum\limits_{i=1}^{N}
\! c_i \! \cdot \! \mathbf{x}^t_i
, ~
{\mathbf{p}^{t}_{+}}
\!=\!
\frac{1}{\|1-C\|_1}
\sum\limits_{i=1}^{N}
\left(1\!- \!c_i\right) \! \cdot \! \mathbf{x}^t_i,
\label{eq:text_prototype}
\end{aligned}
\end{equation}
where $\mathbf{p}^{t}_{-}$ and $\mathbf{p}^{t}_{+}$ are obtained by averaging the textual representations of the inferred anomalous and normal nodes, respectively, thereby capturing the representative semantic patterns of the two groups. 

However, constructing the prototypes alone does not explicitly constrain the positions of individual nodes in the representation space. To further enhance the separation between anomalous and normal textual patterns, we introduce a prototype alignment objective:
\begin{equation}
\begin{aligned}
\mathcal{L}_{\mathrm{align}}
\!&=\!
\frac{1}{N}
\Bigg[
\sum\limits_{i=1}^{N}
c_i
\log
\left(
1+\!
\exp
\left(
\phi(\mathbf{x}^t_i,\mathbf{p}^{t}_{+})
-
\phi(\mathbf{x}^t_i,\mathbf{p}^{t}_{-})
\right)
\right)
\\
&+
(1-c_i)
\log
\left(
1+\!
\exp
\left(
\phi(\mathbf{x}^t_i,\mathbf{p}^{t}_{-})
-
\phi(\mathbf{x}^t_i,\mathbf{p}^{t}_{+})
\right)
\right)
\Bigg],
\label{eq:loss_align}
\end{aligned}
\end{equation}
where $\phi(\cdot,\cdot)$ denotes cosine similarity. For $c_i=1$, the objective pulls $\mathbf{x}^t_i$ toward $\mathbf{p}^{t}_{-}$ and pushes it away from $\mathbf{p}^{t}_{+}$, while for $c_i=0$, it pulls $\mathbf{x}^t_i$ toward $\mathbf{p}^{t}_{+}$ and pushes it away from $\mathbf{p}^{t}_{-}$. This objective yields a more discriminative textual representation space with a clearer normal-anomalous boundary.

The probability estimation and prototype alignment objectives optimize different components of the textual module:
\begin{equation}
    \mathcal{L}_{prob} \stackrel{\mathtt{update}}{\longrightarrow } f^{\Theta}_{prob}(\cdot),~
    \mathcal{L}_{align} \stackrel{\mathtt{update}}{\longrightarrow } ~f^{\Theta}_{txt}(\cdot)~,
\label{eq:upgrade}
\end{equation}
where $\mathcal{L}_{\mathrm{prob}}$ optimizes the text anomaly
probability estimator to estimate textual anomaly probabilities, while $\mathcal{L}_{\mathrm{align}}$ optimizes the text encoder to learn discriminative semantic representations of anomalies.

After training, the textual anomaly score of node $v_i$ is defined as
\begin{equation}
\begin{aligned}
\mathcal{S}^{t}(v_i)\!
=\!\!
f^{\Theta}_{prob}
(\mathbf x^t_i) 
\!+\!
\phi\left(
\mathbf x^t_i,
\mathbf p^{t}_-
\right)
\!+\!
\|
\mathbf{x^t}_i
\!-\!
\frac{1}{|\mathcal N(i)|}
\!\!\sum_{j\in\mathcal N(i)}\!\!\!
\mathbf{x}^t_j
\|_2^2,
\label{eq:text_score}
\end{aligned}
\end{equation}
where the first term is the textual anomaly probability estimated by
$f_{\mathrm{prob}}^{\Theta}$, while the second measures the semantic affinity between the node representation and the textual anomaly prototype learned from the source graphs. The third term captures the deviation between a node's textual representation and its local textual neighborhood. Before summation, all score terms are standardized using Z-score normalization to eliminate their scale differences. By integrating these standardized signals, $\mathcal{S}^{t}(v_i)$ provides textual evidence for the final decoupled anomaly scoring.

\subsection{Topological Prototype Learning}
Following ProMoS~\cite{xu2026generalist}, we learn two node representations to model transferable structural normality. A Graph Transformer is first trained with self-supervised contrastive learning to obtain topology-aware node representations:
\begin{equation}
\begin{aligned}
     \mathbf{H}=&f^{\Theta}_{\mathtt{GNN}}(\mathbf{X},\mathbf{A}) = \mathtt{Graph~Transformer} (\mathbf{X},\mathbf{A}),
\label{eq:graph_encoder}
\end{aligned}
\end{equation}
where $\mathbf{H}=[\mathbf{h}_1,\ldots,\mathbf{h}_N]^\top$ encodes the node attributes together with their topological contexts.

To obtain a complementary representation, we further employ an MLP that takes only the node features as input and learns to approximate the topology-aware representations through knowledge distillation:
\begin{equation}
\begin{aligned}
 \mathbf{H}'=&f^{\Theta}_{\mathtt{MLP}} \left(\mathbf{X})= \sigma((\sigma(\mathbf{X}\mathbf{W} +\mathbf{b}))\mathbf{W}' + \mathbf{b}'\right)
\label{eq:MLP}
\end{aligned}
\end{equation}

Although the MLP does not directly use the adjacency matrix, the distillation process transfers topology-related knowledge from $\mathbf{H}$ to $\mathbf{H}'$. The two representations therefore provide complementary views for subsequent topological prototype construction.

To capture diverse structural normality patterns, we apply $K$-means clustering to the topology-aware representations $\mathbf{H}$ and use the resulting cluster centers as topological prototypes:
\begin{equation}
    {\mathbf{P}^{s}} = \mathtt{K}\text{-}\mathtt{means}(\mathbf{H}, K) = \{\mathbf{p}^s_1, \ldots, \mathbf{p}^s_K\},
\label{eq:struct_prototype}
\end{equation}
where each $\mathbf{p}^s_k$ represents a typical structural normality pattern in the topology-aware representation space.
To ensure that the projected representation $\mathbf{h}'_i$ preserves the structural patterns encoded in $\mathbf{h}_i$, we align their similarity distributions over the shared topological prototypes. Specifically, the structural consistency objective is defined as
\begin{equation}
\mathcal{L}_{str}\! = \!\!\frac{1}{N}\!\! \sum_{i=1}^{N}\! \mathtt{KL}\!
\left(
\operatorname{softmax}\!
\left(
\mathbf{h}_i \!\cdot\! (\mathbf{P}^{s})^\top
\right)\!\! \parallel \!
\operatorname{softmax}\!
\left(
\mathbf{h}'_i\!\cdot\!(\mathbf{P}^{s})^\top 
\right)\!
\right)
\label{eq:loss_str}
\end{equation}
where $\mathtt{KL}(\cdot\parallel\cdot)$ denotes the Kullback--Leibler divergence. This objective encourages $\mathbf{h}'_i$ to retain the relative affinities of $\mathbf{h}_i$ to different topological prototypes, thereby preserving its topology-aware structural semantics.
Based on these prototypes and the two complementary node representations, we define the topological anomaly score of node $v_i$ as
\begin{equation}
\begin{aligned}
\mathcal{S}^{s}(v_i)\!
=&\mathtt{KL}
\left(
\operatorname{softmax}
\left(
\mathbf{h}_i\!\cdot\!(\mathbf{P}^{s})^\top
\right)\!\parallel
\operatorname{softmax}
\left(
\mathbf{h}'_i\cdot(\mathbf{P}^{s})^\top
\right)\!
\right)
\\
&+    \left\|
\mathbf{h}_i-\mathbf{p}_m^s
\right\|_2^2,~m=\arg\min_{k}
\left\|
\mathbf{h}_i-\mathbf{p}_k^s
\right\|_2^2.
\label{eq:struct_score}
\end{aligned}
\end{equation}
Specifically, the two terms measure the discrepancy between the similarity distributions of the two node representations over the topological prototypes and the deviation from the nearest topological prototype, respectively. Before summation, both score terms are standardized using Z-score normalization to eliminate scale differences.

\subsection{Decoupled Dual-Prototype Anomaly Scoring}
The textual and topological branches model semantic and structural anomalies in separate prototype spaces. The two branches produce the textual anomaly score and the topological anomaly score, respectively. ProTAGAD combines the two scores only at the final scoring stage, thereby avoiding the interference caused by deeply coupled representations. The overall anomaly score of node $v_i$ is defined as
\begin{equation}
\begin{aligned}
\mathcal{S}(v_i)
=
{\mathcal{S}^{t}(v_i)}
+
{\mathcal{S}^{s}(v_i)}.
\label{final_score}
\end{aligned}
\end{equation}
Before final aggregation, the textual and topological anomaly scores are independently standardized using Z-score normalization to ensure comparable scales. Consequently, a node is considered more anomalous when it exhibits a large standardized deviation in either textual semantics or topological structure, resulting in a higher final anomaly score $\mathcal{S}(v_i)$.

\section{Experiments}
\subsection{Experimental Settings}

\subsubsection{Datasets.}

We adopt a cross-domain source/target split across 14 text-attributed graphs spanning the citation, e-commerce, web, and encyclopedia domains~\cite{sen2008collective,hu2020open,yan2023comprehensive,yan2024graph,mernyei2020wikics,wang2025generalization}. We synthesize and inject realistic anomalies that mirror real-world scenarios, including off-topic papers, citation manipulation, misleading products, fraudulent co-purchases, fake encyclopedia entries, and promotional content. Building on CMUCL~\cite{xu2025text}, we further introduce contextual and structural anomalies to construct a challenging benchmark to evaluate zero-shot cross-domain generalization.

\begin{table}[t]
\centering
\small  
\setlength{\tabcolsep}{2.5pt}  
\renewcommand{\arraystretch}{0.9}
\setlength{\aboverulesep}{1pt}
\setlength{\belowrulesep}{1pt}
\begin{tabular}{lcccccc}
\toprule
Dataset & Train & Test & Nodes & Edges & AvgLen & Anomaly \\
\midrule
\multicolumn{7}{c}{Citation network} \\
\midrule
Cora     & $\checkmark$ & --    & 2,791   & 10,990    & 135.45 & 164 \\
Citeseer & --           & $\checkmark$ & 3,262   & 3,684     & 153.94 & 192 \\
Pubmed   & --           & $\checkmark$ & 21,248  & 92,154    & 256.08 & 1,184 \\
Arxiv    & $\checkmark$ & --    & 176,514 & 1,225,722 & 179.70 & 10,162 \\
\midrule
\multicolumn{7}{c}{E-commerce network} \\
\midrule
History  & $\checkmark$ & --    & 43,503  & 373,468   & 228.36 & 2,494 \\
Children & $\checkmark$ & --    & 85,137  & 1,591,763 & 209.12 & 4,612 \\
Grocery  & --           & $\checkmark$ & 17,908  & 147,332   & 67.36  & 1,024 \\
Movies   & --           & $\checkmark$ & 17,625  & 166,690   & 81.85  & 1,000 \\
Toys     & --           & $\checkmark$ & 21,283  & 117,424   & 74.50  & 1,240 \\
Fitness  & --           & $\checkmark$ & 187,918 & 3,113,588 & 21.87  & 10,384 \\
Products & $\checkmark$ & --    & 46,617  & 117,064   & 110.65 & 2,752 \\
\midrule
\multicolumn{7}{c}{Web network} \\
\midrule
Cornell & --           & $\checkmark$ & 197    & 596     & 261.27 & 21 \\
Texas   & --           & $\checkmark$ & 192    & 608     & 192.47 & 21 \\
WikiCS  & $\checkmark$ & --    & 14,485 & 443,558 & 422.00 & 704 \\
\bottomrule
\end{tabular}
\caption{Statistics of datasets.}
\label{tab:dataset_statistics}
\end{table}

\subsubsection{Baselines.}
We compare ProTAGAD with 18 representative baselines: 
(1) \textit{GAD methods}---
DOMINANT~\cite{ding2019deep},
BGNN~\cite{ivanov2021boost},
BWGNN~\cite{tang2022rethinking}, GHRN~\cite{gao2023addressing},
CoLA~\cite{liu2021anomaly}, 
HCM-A~\cite{huang2022hop}, TAM~\cite{qiao2023truncated}, 
Semi-GGAD~\cite{qiao2024generative}, CAGAD~\cite{xiao2024counterfactual} and GCTAM~\cite{ijcai2025p405};  
(2) \textit{GGAD methods}---ARC~\cite{liu2024arc}, 
IA-GGAD~\cite{zhang2026ia},
AnomalyGFM~\cite{qiao2025anomalygfm}, UNPrompt~\cite{niu2024zero}, 
OWLEYE~\cite{zheng2026owleye} and 
ProMoS~\cite{xu2026generalist}; 
(3) \textit{TAGAD methods}---CMUCL~\cite{xu2025text} 
and CoLL~\cite{xu2025court}.

\subsubsection{Implementation.}
We report AUROC and AUPRC as mean $\pm$ standard deviation over five random seeds~\cite{tang2023gadbench}. Each method is trained once on $\mathcal{T}_{\mathrm{train}}$ and directly evaluated on $\mathcal{T}_{\mathrm{test}}$ under a pretrain-only protocol. We use BGE\cite{xiao2024c}, GraphTransformer, and DeepSeek4-Flash as the text encoder, graph encoder, and LLM agent, respectively. The agent is used only for offline preprocessing, with its cached outputs shared across all seeds; no LLM calls are required during training or inference. This preprocessing takes approximately $395.8$ seconds and costs USD~$6.64$. Additional implementation details are provided in Appendix~B.


\subsection{Main Results}

\begin{table*}[t]
\centering
\setlength{\tabcolsep}{1mm}
{\small
\begin{tabular}{c|cccccccc|c}
\toprule
\textbf{Method}
& \textbf{Citeseer}
& \textbf{Pubmed}
& \textbf{Texas}
& \textbf{Grocery}
& \textbf{Movies}
& \textbf{Toys}
& \textbf{Fitness}
& \textbf{Cornell}
& \textbf{Rank} \\
\midrule

\multicolumn{10}{c}{GAD Methods} \\
\midrule
DOMINANT~(2019)
& 54.18$\pm$1.11
& 56.24$\pm$1.96
& 67.96$\pm$1.19
& 55.89$\pm$1.01
& 55.87$\pm$1.07
& 58.38$\pm$0.92
& 67.66$\pm$0.78
& 66.77$\pm$0.87
& 10.12 \\

BGNN*~(2021)
& 55.23$\pm$5.72
& 43.56$\pm$12.80
& 53.05$\pm$7.14
& 53.19$\pm$2.30
& 56.27$\pm$4.00
& 51.94$\pm$2.43
& 50.51$\pm$2.17
& 57.74$\pm$9.96
& 14.00 \\

BWGNN*~(2022)
& 61.26$\pm$1.82
& 54.16$\pm$4.11
& 49.60$\pm$7.91
& 59.20$\pm$1.12
& 55.35$\pm$0.97
& 57.82$\pm$0.77
& 49.76$\pm$1.53
& 50.28$\pm$3.46
& 12.50 \\

GHRN*~(2023)
& 59.34$\pm$1.56
& 50.96$\pm$4.61
& 50.71$\pm$6.07
& 58.65$\pm$2.44
& 52.97$\pm$2.31
& 56.98$\pm$2.45
& 49.69$\pm$2.49
& 56.28$\pm$2.60
& 13.00 \\

CoLA~(2021)
& 57.56$\pm$1.91
& 48.91$\pm$1.10
& 47.86$\pm$6.40
& 53.22$\pm$0.95
& 52.34$\pm$0.81
& 55.90$\pm$0.77
& 54.27$\pm$0.44
& 47.16$\pm$4.27
& 15.00 \\

HCM-A~(2022)
& 60.68$\pm$0.75
& 60.13$\pm$1.18
& 57.57$\pm$1.77
& 64.68$\pm$0.46
& 64.93$\pm$0.36
& 66.55$\pm$0.39
& 65.11$\pm$0.79
& 67.16$\pm$1.64
& 8.00 \\

TAM~(2023)
& 28.01$\pm$0.02
& 74.90$\pm$0.24
& 59.47$\pm$1.18
& 70.67$\pm$0.20
& 66.77$\pm$0.13
& \underline{74.73$\pm$0.13}
& OOM
& 61.07$\pm$2.46
& 8.56 \\

Semi-GGAD~(2024)
& 46.51$\pm$14.23
& 35.87$\pm$6.46
& 42.62$\pm$6.75
& 42.81$\pm$7.73
& 42.93$\pm$6.33
& 44.74$\pm$6.77
& 43.08$\pm$7.35
& 36.75$\pm$5.28
& 18.62 \\

CAGAD~(2024)
& 59.56$\pm$4.00
& 42.11$\pm$6.44
& 59.69$\pm$6.49
& 55.67$\pm$1.61
& 56.32$\pm$3.27
& 56.18$\pm$2.60
& 48.14$\pm$6.08
& 54.41$\pm$2.69
& 12.88 \\

GCTAM~(2025)
& 49.60$\pm$1.11
& 70.76$\pm$0.84
& 67.30$\pm$2.50
& 65.50$\pm$0.56
& 64.34$\pm$0.26
& 68.64$\pm$0.54
& OOM
& 63.93$\pm$2.29
& 9.56 \\

\midrule
\multicolumn{10}{c}{GGAD Methods} \\
\midrule
ARC*~(2024)
& \underline{73.23$\pm$0.01}
& \underline{75.22$\pm$1.03}
& 75.66$\pm$1.07
& \underline{70.95$\pm$0.36}
& 68.16$\pm$0.74
& 74.22$\pm$1.01
& 77.93$\pm$0.61
& \textbf{81.25$\pm$1.94}
& \underline{2.50} \\

IA-GGAD*~(2025)
& 62.29$\pm$1.38
& 60.11$\pm$1.92
& 55.27$\pm$3.55
& 66.42$\pm$0.41
& \underline{68.67$\pm$0.87}
& 69.85$\pm$0.55
& \underline{78.14$\pm$0.50}
& 56.84$\pm$2.45
& 6.62 \\

AnomalyGFM*~(2025)
& 56.19$\pm$2.18
& 68.59$\pm$2.14
& 58.84$\pm$2.38
& 55.93$\pm$2.01
& 51.49$\pm$1.68
& 56.21$\pm$1.14
& 59.97$\pm$1.49
& 74.50$\pm$1.93
& 10.75 \\

UNPrompt*~(2025)
& 65.23$\pm$1.53
& 66.99$\pm$6.99
& 46.80$\pm$5.72
& 54.43$\pm$1.93
& 55.56$\pm$1.86
& 53.56$\pm$1.63
& 66.86$\pm$2.35
& 54.94$\pm$1.62
& 11.75 \\

OWLEYE*~(2026)
& 73.02$\pm$0.20
& 71.32$\pm$0.03
& \underline{77.54$\pm$0.06}
& 66.66$\pm$0.01
& 68.24$\pm$0.01
& 72.82$\pm$0.01
& 72.82$\pm$0.01
& 74.83$\pm$0.06
& 3.50 \\

ProMoS~(2026)
& 64.28$\pm$1.47
& 70.06$\pm$0.58
& 74.88$\pm$1.03
& 65.73$\pm$0.75
& 65.77$\pm$0.39
& 69.34$\pm$0.26
& 77.24$\pm$0.33
& 74.76$\pm$1.20
& 5.12 \\

\midrule
\multicolumn{10}{c}{TAGAD Methods} \\
\midrule
CMUCL~(2025)
& 57.58$\pm$1.28
& 53.13$\pm$1.79
& 51.86$\pm$6.57
& 49.63$\pm$1.56
& 50.69$\pm$0.43
& 48.17$\pm$0.29
& 49.92$\pm$0.70
& 62.41$\pm$3.74
& 14.25 \\

CoLL~(2025)
& 55.79$\pm$1.91
& 56.26$\pm$2.33
& 45.55$\pm$2.49
& 57.32$\pm$2.09
& 59.80$\pm$2.46
& 58.94$\pm$0.86
& 60.02$\pm$1.65
& 51.20$\pm$2.83
& 12.12 \\

\method~
& \textbf{81.88$\pm$0.38}
& \textbf{83.39$\pm$0.42}
& \textbf{81.81$\pm$1.09}
& \textbf{74.85$\pm$0.17}
& \textbf{72.08$\pm$0.33}
& \textbf{77.39$\pm$0.53}
& \textbf{79.37$\pm$0.09}
& \underline{80.31$\pm$0.84}
& \textbf{1.12} \\

$\Delta$
& $\uparrow$\,8.65
& $\uparrow$\,8.17
& $\uparrow$\,4.27
& $\uparrow$\,3.90
& $\uparrow$\,3.41
& $\uparrow$\,2.66
& $\uparrow$\,1.23
& $\downarrow$\,0.94
& -- \\

\bottomrule
\end{tabular}
}
\caption{Performance comparison on different target domains (AUROC, \%, mean $\pm$ std). ``Rank'' is the average rank over eight targets; $\Delta$ is the AUROC difference between \method~and the strongest baseline. Methods marked with * are supervised methods. The best and second-best results are shown in bold and underlined, respectively. OOM denotes out of memory.}
\label{tab:main_results_auc}
\end{table*}

Table~\ref{tab:main_results_auc} summarizes the AUROC results on eight unseen target graphs. ProTAGAD achieves the best performance on seven datasets and attains the lowest average rank of 1.12, demonstrating stable zero-shot cross-domain generalization. The most substantial improvements are observed on Citeseer (+8.65\%) and Pubmed (+8.17\%), indicating that the textual prototype bank effectively preserves fine-grained semantic anomalies and prevents them from being weakened during topological aggregation. The consistent gains on Texas, Grocery, Movies, Toys, and Fitness further suggest that the decoupled topological prototype bank can capture transferable structural normality across different domains. The only exception is Cornell, where ProTAGAD achieves an AUROC of 80.31\%, which is 0.94\% lower than the best baseline. This small gap may result from the extremely limited graph size, which provides insufficiently diverse patterns for stable prototype estimation.
Overall, the results show that independently modeling semantic consistency and topological normality enables ProTAGAD to preserve more discriminative anomaly cues under domain shifts. In addition, the standard deviations remain below 1.1\% on all target graphs, confirming the model's stable performance across random seeds. Full AUPRC results appear in Appendix~D.

\subsection{Ablation Study}
To disentangle the contribution of each prototype module, we evaluate four variants: (i) \textbf{Backbone}, which removes both prototype modules; (ii) \textbf{+ SP}, which incorporates only the structural prototype module; (iii) \textbf{+ TP}, which incorporates only the textual prototype module; and (iv) \textbf{Ours}, which combines both modules. 
\begin{table}[t]
\centering
\small
\setlength{\tabcolsep}{3pt}
\begin{tabular}{lcccc}
\toprule
\textbf{Datasets} & \textbf{Backbone} & \textbf{+ SP} & \textbf{+ TP} & \textbf{Ours} \\
\midrule
Citeseer & 56.53$\pm$2.51 & 76.75$\pm$2.23 & 60.69$\pm$1.08 & \textbf{81.88$\pm$0.38} \\
Pubmed   & 63.14$\pm$4.10 & 75.35$\pm$0.79 & 75.03$\pm$0.57 & \textbf{83.39$\pm$0.42} \\
Grocery  & 49.63$\pm$1.67 & 71.19$\pm$0.39 & 56.85$\pm$1.08 & \textbf{74.85$\pm$0.17} \\
Movies   & 52.96$\pm$2.03 & 68.93$\pm$0.18 & 54.11$\pm$1.64 & \textbf{72.08$\pm$0.33} \\
Toys     & 55.68$\pm$1.67 & 74.57$\pm$0.36 & 58.80$\pm$1.40 & \textbf{77.39$\pm$0.53} \\
Fitness  & 44.03$\pm$2.53 & 77.85$\pm$0.31 & 53.36$\pm$1.06 & \textbf{79.37$\pm$0.09} \\
Cornell  & 54.50$\pm$5.31 & 79.94$\pm$0.74 & 52.91$\pm$3.42 & \textbf{80.31$\pm$0.84} \\
Texas    & 54.69$\pm$2.13 & 78.71$\pm$0.64 & 52.88$\pm$2.72 & \textbf{81.81$\pm$1.09} \\
\bottomrule
\end{tabular}
\caption{Ablation study on different target domains in terms of AUROC (\%).}
\label{tab:ablation}
\end{table}
As shown in Table~\ref{tab:ablation}, structural prototypes consistently improve the Backbone across all target graphs, with particularly large gains on Fitness (+33.82\%), Cornell (+25.44\%), and Texas (+24.02\%), demonstrating their ability to capture transferable structural normality. Textual prototypes also yield notable improvements on Pubmed (+11.89\%) and Fitness (+9.33\%), indicating their effectiveness in capturing semantic anomaly cues. Combining both modules achieves the best performance on every target graph, confirming that textual and structural prototypes provide complementary evidence for cross-domain anomaly detection.

\subsection{Effect of Decoupled Prototype Modeling}

To evaluate decoupled prototype modeling, we compare ProTAGAD with a coupled variant under identical settings. In the coupled variant, the textual features and original node features are first projected to the same dimensionality, separately $\ell_2$-normalized, and then averaged. The fused features replace the original node features as the input to the graph encoder. Both textual and topological prototypes are constructed from the resulting fused representations, and anomaly scores are computed in this shared prototype space.

We further introduce Anomaly Boundary Separability ($\mathcal{ABS}$) to quantify the \textit{BAB} issue. Let $\hat{f}^{\mathrm{+}}(s)$ and $\hat{f}^{\mathrm{-}}(s)$ denote the kernel density estimates of the anomaly scores for normal and anomalous nodes, respectively. Both densities are estimated using a Gaussian kernel with a shared bandwidth determined by applying Scott’s rule to the pooled scores of the two groups. We define
\begin{equation}
\mathcal{ABS}=\!
\sqrt{
\frac{1}{2}
D_{\mathrm{KL}}
\left(
\hat{f}^{\mathrm{+}}(s)\!
\parallel \!M
\right)
+
\frac{1}{2}
D_{\mathrm{KL}}
\left(
\hat{f}^{\mathrm{-}}(s)\!
\parallel \!M
\right)
}
\end{equation}
where $M=\frac{1}{2}(\hat{f}^{\mathrm{+}}+\hat{f}^{\mathrm{-}})$ denotes the mixture distribution, and $D_{\mathrm{KL}}(\cdot\parallel\cdot)$ represents the Kullback--Leibler divergence. A larger $\mathcal{ABS}$ indicates clearer separation between normal and anomalous nodes and thus a less severe \textit{BAB} issue.

As shown in Table~\ref{tab:coupled_decoupled}, ProTAGAD consistently outperforms the coupled variant on all eight target graphs, increasing the average AUROC from 61.36\% to 78.89\% and the average $\mathcal{ABS}$ from 0.2260 to 0.4553. The lower $\mathcal{ABS}$ of the coupled variant indicates that normal and anomalous nodes exhibit more similar anomaly-score distributions, leading to weaker boundary separability and a more severe \textit{BAB} issue. 
In contrast, decoupled prototype modeling preserves the complementary anomaly information of the two modalities, resulting in more discriminative anomaly scores and clearer normal--anomalous boundaries.


\begin{table}[h]
\centering
\footnotesize
\setlength{\tabcolsep}{1.6pt}
\begin{tabular}{l|ccc|ccc}
\toprule
& \multicolumn{3}{c|}{\textbf{AUROC (\%) $\uparrow$}}
& \multicolumn{3}{c}{\textbf{$\mathcal{ABS}$ $\uparrow$}} \\
\cmidrule(lr){2-4}
\cmidrule(lr){5-7}
\textbf{Dataset}
& \textbf{Coupled}
& \textbf{Decoupled}
& \textbf{$\Delta$}
& \textbf{Coupled}
& \textbf{Decoupled}
& \textbf{$\Delta$} \\
\midrule
Citeseer
& 74.51
& \textbf{81.88}
& 7.37
& 0.4529
& \textbf{0.5416}
& 0.0887 \\

Pubmed
& 62.40
& \textbf{83.39}
& 20.99
& 0.1876
& \textbf{0.5336}
& 0.3460 \\

Grocery
& 58.12
& \textbf{74.85}
& 16.73
& 0.1227
& \textbf{0.3861}
& 0.2634 \\

Movies
& 60.09
& \textbf{72.08}
& 11.99
& 0.1729
& \textbf{0.3536}
& 0.1807 \\

Toys
& 59.73
& \textbf{77.39}
& 17.66
& 0.1523
& \textbf{0.4253}
& 0.2730 \\

Fitness
& 62.28
& \textbf{79.37}
& 17.09
& 0.1723
& \textbf{0.4948}
& 0.3225 \\

Cornell
& 59.65
& \textbf{80.31}
& 20.66
& 0.3043
& \textbf{0.4369}
& 0.1326 \\

Texas
& 54.13
& \textbf{81.81}
& 27.68
& 0.2428
& \textbf{0.4709}
& 0.2281 \\
\bottomrule
\end{tabular}
\caption{Comparison between coupled and decoupled variants in terms of AUROC (\%) and anomaly boundary separability ($\mathcal{ABS}$). Higher values indicate better detection performance and clearer anomaly boundaries.}
\label{tab:coupled_decoupled}
\end{table}

\subsection{Parameter Sensitivity}
\begin{figure}[h]
    \centering
    \begin{minipage}{0.49\linewidth}
        \centering
        \includegraphics[width=\linewidth]{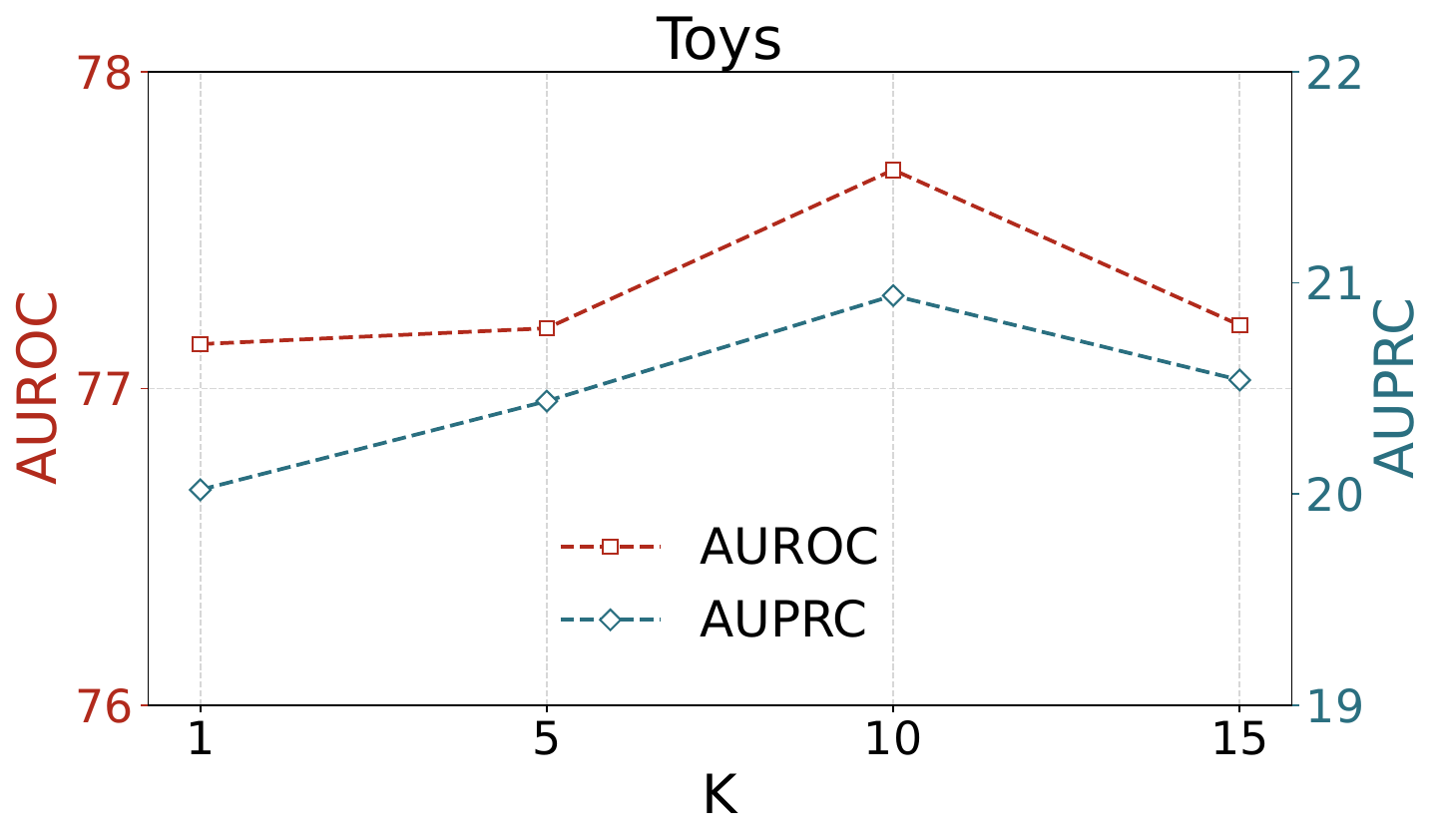}
        \small (a) Toys
    \end{minipage}
    \hspace{0pt}
    \begin{minipage}{0.49\linewidth}
        \centering
        \includegraphics[width=\linewidth]{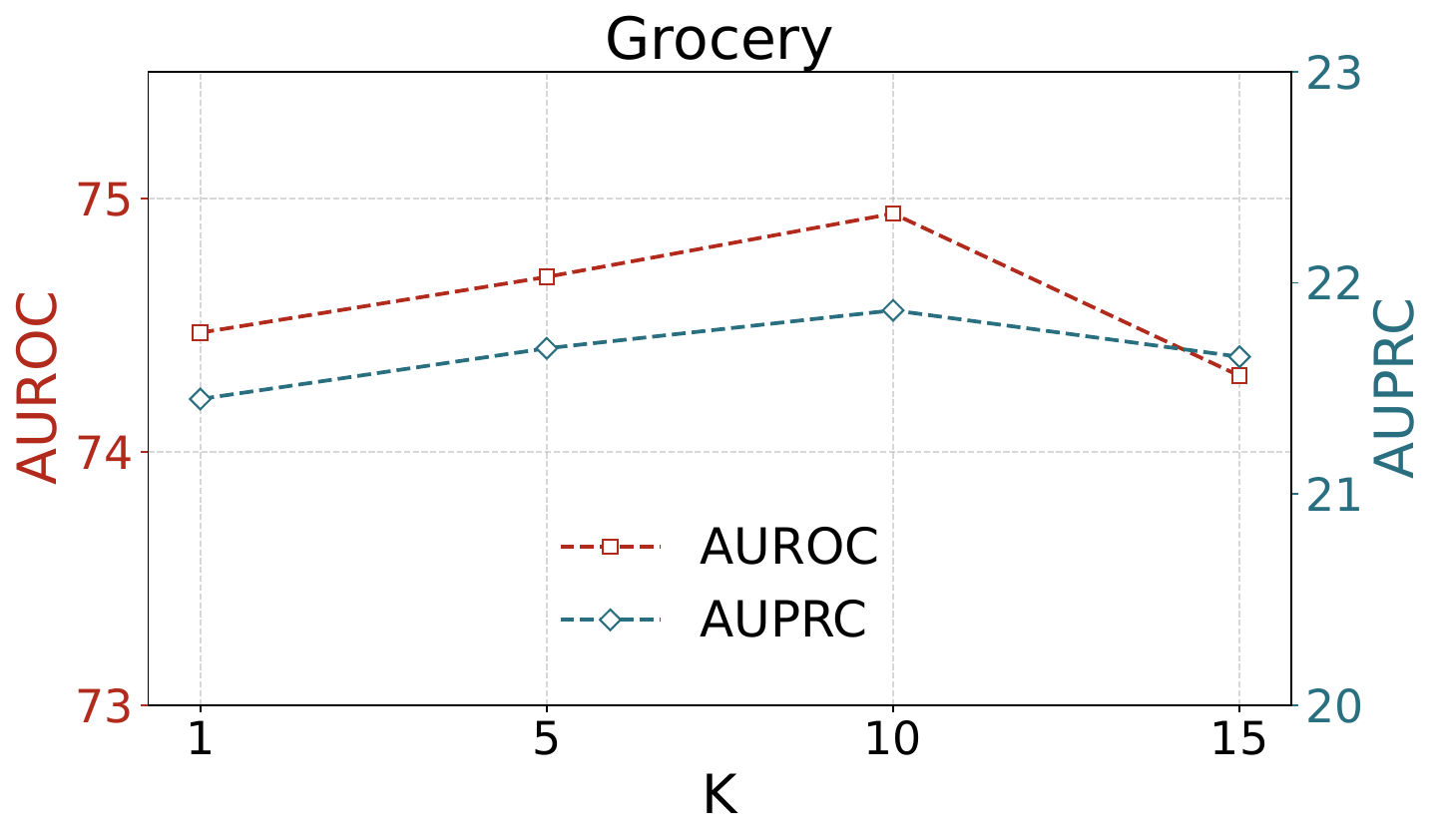}
        \small (b) Grocery
    \end{minipage}
    \caption{Impact of K for \method~on Toys and Grocery datasets.}
    \label{fig:k_sensitivity}
\end{figure}

We investigate the effect of the number of structural prototypes $K$ in Eq.~(\ref{eq:struct_prototype}) on the Toys and Grocery datasets. As shown in Figure~\ref{fig:k_sensitivity}, both AUROC and AUPRC gradually improve as $K$ increases from 1 to 10, and achieve their best performance at $K=10$. A small $K$ provides insufficient prototypes to characterize the diverse structural normality patterns across graphs. In contrast, an excessively large $K$ may fragment the normal structural distribution and introduce domain-specific noise, leading to performance degradation when $K=15$. Overall, ProTAGAD remains relatively stable under different values of $K$, while $K=10$ provides the best balance between prototype diversity and cross-domain generalization.

\subsection{Efficiency Analysis}
\begin{figure}[h]
    \centering
    \begin{minipage}{0.48\columnwidth}
        \centering
        \includegraphics[width=\linewidth]{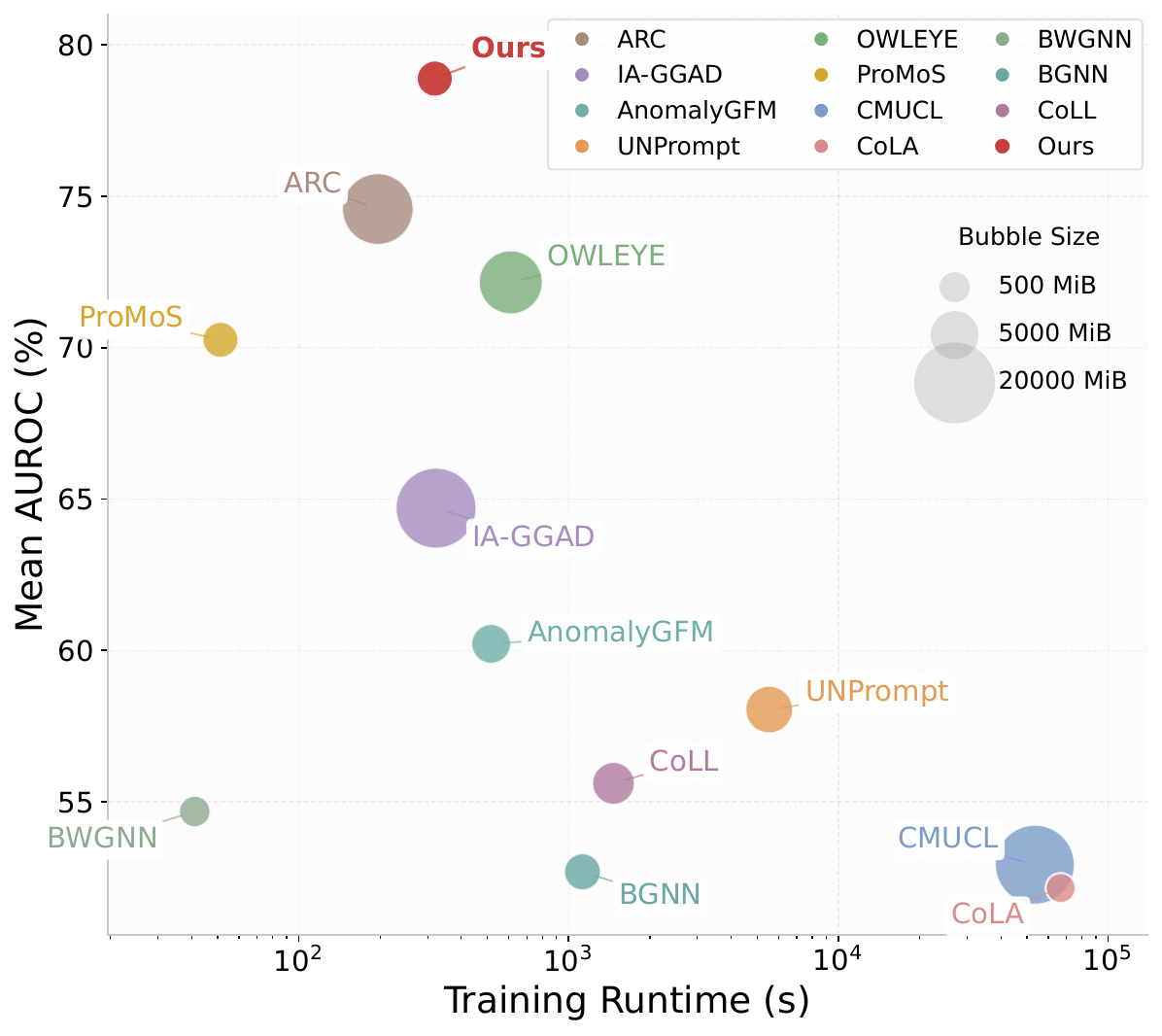}
        \par\smallskip
        (a) Training
    \end{minipage}
    \hfill
    \begin{minipage}{0.48\columnwidth}
        \centering
        \includegraphics[width=\linewidth]{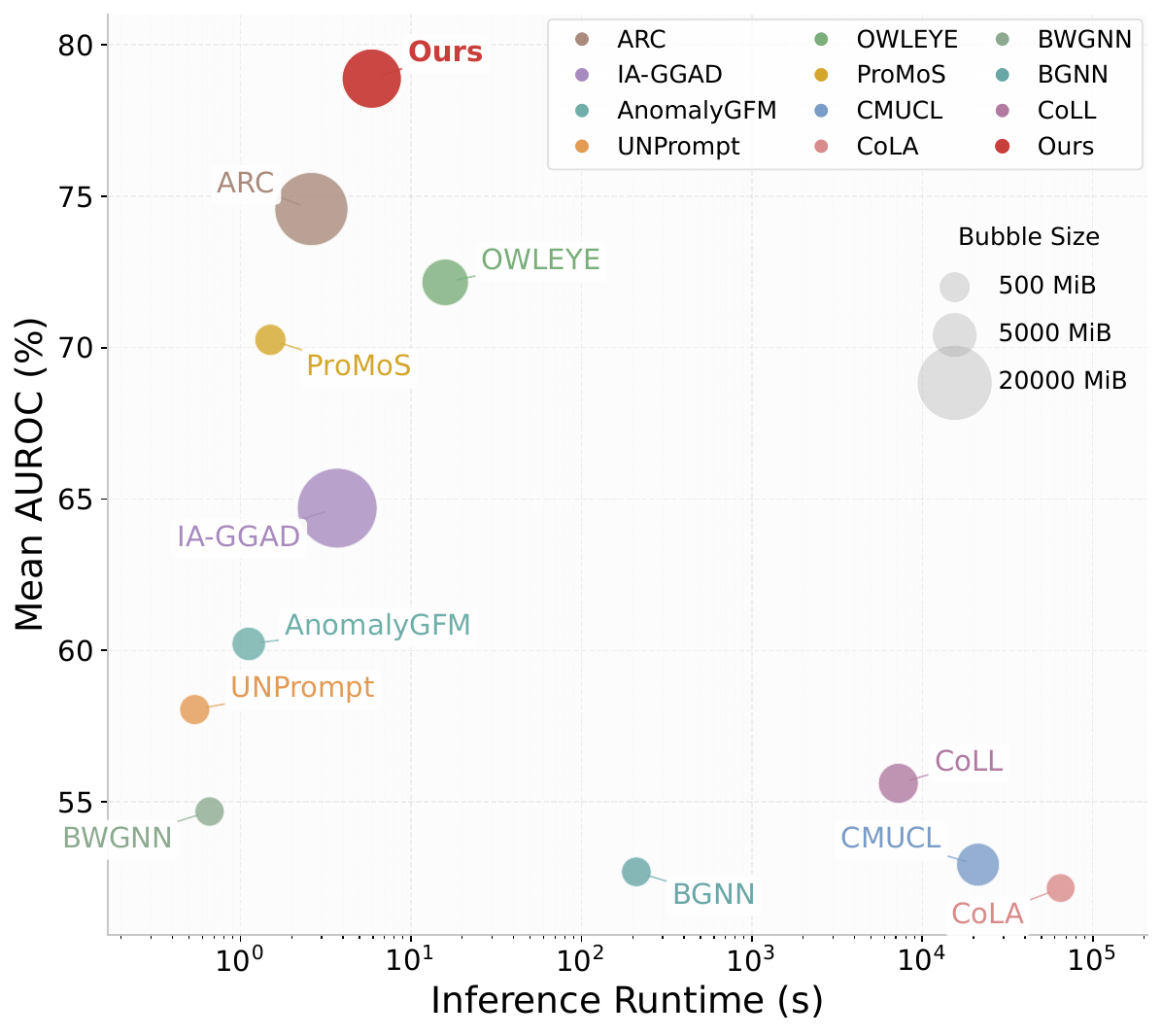}
        \par\smallskip
        (b) Inference
    \end{minipage}
    \caption{Efficiency comparison of different methods in terms of runtime and GPU memory usage during training and inference.}
    \label{fig:efficiency}
\end{figure}

Figure~\ref{fig:efficiency} compares the runtime, GPU memory usage, and mean AUROC of different methods during training and inference, where bubble size denotes memory consumption. ProTAGAD lies near the upper-left region in both panels, indicating that it achieves the highest detection performance with competitive computational efficiency. It also maintains relatively low GPU memory usage during training, while its inference memory consumption is comparatively high. Overall, ProTAGAD achieves a favorable trade-off between detection performance and computational efficiency.

\section{Conclusion}
We presented \textbf{ProTAGAD}, a zero-shot foundation model for generalist text-attributed graph anomaly detection on unseen graphs. By decoupling textual and topological representation learning and separately learning textual and structural prototypes, ProTAGAD alleviates the \textit{Blurred-Anomaly-Boundary} (\textit{BAB}) issue. Across 14 TAG datasets, ProTAGAD achieves the best AUROC on seven of eight target graphs with an average rank of 1.12. Ablation studies and coupled-versus-decoupled comparisons further validate the complementarity of the two prototype branches and the effectiveness of the proposed design. 
Future work will explore more memory-efficient prototype learning and extend ProTAGAD to large-scale graph domains.

\bibliography{references}

\clearpage
\appendix

\section{Description of Datasets}
We evaluate ProTAGAD on 14 text-attributed graph datasets spanning three domains: citation networks, e-commerce networks, and web networks. Following the cross-domain evaluation setting, Cora, Arxiv, Children, Products, History, and WikiCS are used as source graphs for training, while Citeseer, Pubmed, Grocery, Movies, Toys, Fitness, Cornell, and Texas serve as unseen target graphs for testing. These datasets exhibit substantial variations in graph scale, topology, textual content, and domain semantics, providing a diverse benchmark for evaluating zero-shot cross-domain generalization. 

\begin{itemize}

    \item \textbf{Cora}~\cite{sen2008collective}, \textbf{Citeseer}, and \textbf{Pubmed} are citation networks in which nodes represent academic papers and edges denote citation relationships. The textual attributes of each node are constructed from the title and abstract of the corresponding paper.
    
    \item \textbf{Arxiv}~\cite{hu2020open} is constructed from the ogbn-arxiv dataset, where nodes correspond to research papers and directed edges indicate citation links. We adopt its raw-text version released by the TAG benchmark~\cite{yan2023comprehensive}, which augments the original citation topology with the title and abstract of each paper.

    \item \textbf{Products}~\cite{hu2020open} is a subset of the ogbn-products graph, where nodes represent commercial products, edges indicate co-purchase relationships, and node attributes contain product titles and descriptions.

    \item \textbf{Children}, \textbf{History}~\cite{yan2023comprehensive}, \textbf{Grocery}, \textbf{Movies}, and \textbf{Toys}~\cite{yan2024graph} are category-specific Amazon product graphs, where nodes represent products, edges capture co-purchase relationships, and product titles and descriptions serve as textual attributes.

    \item \textbf{Fitness}~\cite{hu2020open,yan2023comprehensive} is a large-scale Amazon product graph, where nodes represent fitness-related products, edges capture co-purchase relationships, and product descriptions serve as textual attributes.

    \item \textbf{WikiCS}~\cite{mernyei2020wikics} is a web network composed of Wikipedia pages related to computer science. Nodes represent Wikipedia entries, and edges correspond to hyperlinks between pages. The textual attributes are derived from entry names and page contents.

    \item \textbf{Cornell}~\cite{wang2025generalization} is a web network collected from the Cornell University website. Nodes represent webpages, and edges denote hyperlinks between them. The node attributes are constructed from the original textual contents of the webpages.

    \item \textbf{Texas}~\cite{wang2025generalization} is a web network collected from the computer science website of the University of Texas. Nodes correspond to webpages, and edges represent hyperlink relationships. The textual attributes consist of webpage titles and body texts.

\end{itemize}

\noindent\textbf{Anomaly Injection.}
For the datasets requiring anomaly injection, we follow CMUCL~\cite{xu2025text} and construct anomalous nodes through an inner-injection stage and an outer-injection stage. The inner-injection stage perturbs the original textual attributes and graph topology to generate two contextual and two structural anomaly types. For each contextual anomaly, we randomly select a normal target node and sample $K=50$ candidate nodes from the same graph. The candidate whose text embedding has the lowest cosine similarity to that of the target node is selected as the anomaly source. Contextual insertion inserts textual content from the source node into an eligible field of the target text, whereas contextual replacement substitutes an existing sentence or text block with source content. Field boundaries, including titles, descriptions, and abstracts, are preserved during both operations. For structural perturbation, dense-clique anomalies are generated by randomly sampling groups of four nodes and fully connecting the nodes within each group. Random-edge anomalies are constructed by connecting a selected target node to randomly sampled nodes, where the number of newly added edges is drawn from the empirical degree distribution of the original graph.

The outer-injection stage further introduces domain-grounded anomalies using external anomaly-source nodes. Their textual content is used for contextual insertion or replacement, while structural anomalies are generated by appending external nodes to the graph and either forming dense four-node cliques with selected target nodes or connecting them to target nodes through randomly added edges. Existing anomaly labels are retained throughout the process, and the target nodes assigned to different anomaly types are kept disjoint. By default, the inner-injection stage introduces approximately $4\%$ anomalous nodes in total, while the outer-injection stage further injects approximately $2\%$ anomalous nodes. The labels produced by the two stages are combined to obtain the final anomaly annotations.

\section{Details of Implementation}

\paragraph{Hyper-parameters.}
We use a shared hyper-parameter configuration for all datasets without target-specific tuning. The aligned feature dimension and hidden dimension are both set to 64, and the 768-dimensional BGE embeddings are projected through a two-layer text adapter. The Graph Transformer contains 2 layers and 4 attention heads with a dropout rate of 0.2. The hidden dimension and dropout rate of the multilayer perceptrons are set to 128 and 0.1, respectively. The number of topological prototypes is selected from $\{1,5,10,15\}$ and set to 10. The warm-up and main training epochs are set to 5 and 3. The feature dropout rates are 0.1 and 0.2, the edge dropout rates are 0.2 and 0.4, and the contrastive temperature is 0.4. At most 4,096 nodes are used for each contrastive loss computation. We use Adam, with the learning rates for the model parameters and prototypes searched over $\{10^{-5},5\times10^{-5},10^{-4},5\times10^{-4},10^{-3},5\times10^{-3},10^{-2},5\times10^{-2}\}$ and both set to $5\times10^{-3}$. The weight decay is set to $5\times10^{-5}$, and the maximum gradient norm is 5.0. For each source graph, we sample 64 subgraphs per epoch with at most 256 nodes and an expansion length of 3. We additionally use 64 subgraphs for prototype initialization and at most 64 skill-guided anchor subgraphs. All score components are normalized using $z$-score normalization.

\paragraph{Baseline Implementation.}
We use the official implementations and recommended hyper-parameter settings whenever they are publicly available. All baselines use the same source--target dataset split, input features, and five random seeds as ProTAGAD. Each method is trained on $\mathcal{T}_{\mathrm{train}}$ and directly evaluated on the unseen graphs in $\mathcal{T}_{\mathrm{test}}$ without target-specific fine-tuning or hyper-parameter selection. We retain the original supervision setting and anomaly-scoring mechanism of each baseline to avoid method-specific modifications that could affect the comparison.

\paragraph{LLM Agent Implementation.}
The LLM agent uses DeepSeek4-Flash through the API provided by the DeepSeek Open Platform and is invoked only during offline preprocessing on the source graphs. For each source graph, approximately $6\%$ of nodes are selected through stratified random sampling over source-graph node groups. For each selected node, the input contains its original textual attribute and the textual context of its one-hop neighborhood, with at most 10 neighboring nodes retained. The center-node text is truncated to at most 5,000 characters, and each neighboring node text is truncated to at most 320 characters. The maximum API concurrency is set to 1,000. All generated outputs are cached and shared across the five random seeds. The complete preprocessing involves 30,996 API requests and consumes $2.003\times10^{8}$ tokens. The entire process takes approximately 395.8 seconds and costs USD~$6.64$. No LLM calls are required during model training or inference.

\paragraph{Implementation Details.}
All experiments were conducted on a server equipped with an Intel Xeon Platinum 8458P CPU and four NVIDIA A40 GPUs, each with 48\,GB of memory. The implementation was developed using Python 3.10.19, PyTorch 2.6.0 with CUDA 11.8, and PyTorch Geometric 2.7.0. We used \texttt{bge-base-en-v1.5} as the text encoder, a Graph Transformer as the graph encoder, and DeepSeek4-Flash as the LLM agent.

\paragraph{Metrics.}
We evaluate anomaly detection performance using AUROC and AUPRC. AUROC measures the
ranking quality of anomaly scores, while AUPRC is more informative under class imbalance.
Higher AUROC and AUPRC indicate better detection performance. All results are reported as
mean$\pm$standard deviation over five independent runs.

\section{Description of Baselines}
To ensure a fair and comprehensive evaluation of ProTAGAD, we compare it against 18 representative baselines spanning three categories: conventional graph anomaly detection (GAD), generalized graph anomaly detection (GGAD), and text-attributed graph anomaly detection (TAGAD). These categories represent common methodological frameworks in graph anomaly detection and include several state-of-the-art (SOTA) models, providing a broad basis for performance comparison under different assumptions and settings.

\subsection{GAD Methods}

\begin{itemize}
    \item \textbf{BGNN}~\cite{ivanov2021boost} integrates gradient-boosted decision trees with graph neural networks to jointly model node attributes and graph structure.

    \item \textbf{BWGNN}~\cite{tang2022rethinking} employs localized Beta-wavelet filters to capture anomalous signals across different graph frequency bands.

    \item \textbf{GHRN}~\cite{gao2023addressing} suppresses potentially heterophilic edges using high-frequency graph signals to improve anomaly discrimination.

    \item \textbf{DOMINANT}~\cite{ding2019deep} jointly reconstructs node attributes and graph structure, using their reconstruction errors to identify anomalous nodes.

    \item \textbf{CoLA}~\cite{liu2021anomaly} performs anomaly detection through contrastive learning between target nodes and their sampled neighborhood subgraphs.

    \item \textbf{HCM-A}~\cite{huang2022hop} adopts hop-count prediction as a self-supervised task and estimates anomalies based on prediction errors and uncertainty.

    \item \textbf{TAM}~\cite{qiao2023truncated} learns anomaly-discriminative representations by maximizing local affinity while truncating suspicious graph interactions.

    \item \textbf{Semi-GGAD}~\cite{qiao2024generative} generates pseudo-anomalous nodes from a small set of labeled normal nodes to train a one-class detector.

    \item \textbf{CAGAD}~\cite{xiao2024counterfactual} uses counterfactual graph augmentation to construct anomaly-aware neighborhoods and improve anomaly separability.

    \item \textbf{GCTAM}~\cite{ijcai2025p405} combines contextual and global affinity with adaptive truncation to reduce incorrect anomaly judgments.
\end{itemize}

\subsection{GGAD Methods}

\begin{itemize}
    \item \textbf{ARC}~\cite{liu2024arc} performs cross-graph anomaly detection through feature alignment, ego-neighbor residual encoding, and in-context reconstruction.

    \item \textbf{AnomalyGFM}~\cite{qiao2025anomalygfm} aligns neighborhood residual representations with graph-independent prototypes for zero-shot and few-shot anomaly detection.

    \item \textbf{UNPrompt}~\cite{niu2024zero} learns transferable neighborhood prompts and detects anomalies according to node-attribute prediction discrepancies.

    \item \textbf{IA-GGAD}~\cite{zhang2026ia} learns anomaly-relevant invariant representations while reducing feature and structural shifts across graphs.

    \item \textbf{OWLEYE}~\cite{zheng2026owleye} captures transferable normal patterns through cross-domain feature alignment and multi-pattern dictionary learning.

    \item \textbf{ProMoS}~\cite{xu2026generalist} employs prototype-guided knowledge distillation and a mixture-of-students architecture for zero-shot anomaly detection.
\end{itemize}

\subsection{TAGAD Methods}

\begin{itemize}
    \item \textbf{CMUCL}~\cite{xu2025text} jointly models textual attributes and graph topology through multi-scale cross-modal and uni-modal contrastive learning.

    \item \textbf{CoLL}~\cite{xu2025court} combines multi-LLM semantic reasoning with GNN-based topology modeling to detect contextual and structural anomalies.
\end{itemize}

\begin{table*}[t]
\centering
\setlength{\tabcolsep}{1mm}
{\small
\begin{tabular}{c|cccccccc|c}
\toprule
\textbf{Method}
& \textbf{Citeseer}
& \textbf{Pubmed}
& \textbf{Texas}
& \textbf{Grocery}
& \textbf{Movies}
& \textbf{Toys}
& \textbf{Fitness}
& \textbf{Cornell}
& \textbf{Rank} \\
\midrule

\multicolumn{10}{c}{GAD Methods} \\
\midrule
DOMINANT~(2019)
& 9.45$\pm$1.32
& 9.37$\pm$0.93
& 21.14$\pm$1.93
& 7.60$\pm$0.23
& 6.97$\pm$0.29
& 8.10$\pm$0.23
& 13.23$\pm$1.62
& 31.16$\pm$3.65
& 10.50 \\

BGNN*~(2021)
& 9.01$\pm$3.31
& 5.85$\pm$3.20
& 12.14$\pm$2.10
& 6.61$\pm$0.65
& 6.99$\pm$0.91
& 6.54$\pm$0.92
& 5.66$\pm$0.53
& 18.69$\pm$6.67
& 14.88 \\

BWGNN*~(2022)
& 20.81$\pm$3.69
& 13.63$\pm$5.85
& 12.93$\pm$2.68
& 9.36$\pm$0.82
& 8.17$\pm$1.27
& 8.28$\pm$0.42
& 8.59$\pm$1.65
& 15.66$\pm$1.35
& 11.12 \\

GHRN*~(2023)
& 18.52$\pm$1.34
& 15.18$\pm$10.03
& 13.45$\pm$2.50
& 9.27$\pm$1.26
& 8.11$\pm$1.34
& 8.18$\pm$0.92
& 9.04$\pm$1.88
& 16.24$\pm$1.59
& 10.62 \\

CoLA~(2021)
& 8.94$\pm$1.42
& 5.43$\pm$0.17
& 13.88$\pm$3.37
& 6.35$\pm$0.21
& 6.04$\pm$0.21
& 7.14$\pm$0.17
& 6.33$\pm$0.13
& 10.28$\pm$0.77
& 16.00 \\

HCM-A~(2022)
& 10.72$\pm$0.67
& 9.41$\pm$0.61
& 13.99$\pm$0.97
& 12.27$\pm$0.67
& 11.14$\pm$0.46
& 15.73$\pm$0.60
& 8.86$\pm$0.50
& 18.89$\pm$2.69
& 9.50 \\

TAM~(2023)
& 3.92$\pm$0.00
& 11.87$\pm$0.22
& 15.93$\pm$0.88
& 10.01$\pm$0.09
& 8.55$\pm$0.05
& 12.64$\pm$0.09
& OOM
& 14.29$\pm$0.93
& 12.19 \\

Semi-GGAD~(2024)
& 5.78$\pm$1.96
& 4.20$\pm$0.53
& 14.49$\pm$7.91
& 4.84$\pm$0.82
& 4.78$\pm$0.70
& 5.16$\pm$0.81
& 4.71$\pm$0.81
& 11.68$\pm$4.61
& 17.62 \\

CAGAD~(2024)
& 17.03$\pm$3.90
& 5.59$\pm$1.58
& 21.59$\pm$7.46
& 7.97$\pm$1.28
& 8.52$\pm$1.38
& 7.37$\pm$0.54
& 7.97$\pm$3.75
& 15.73$\pm$2.27
& 11.62 \\

GCTAM~(2025)
& 7.69$\pm$0.23
& 26.77$\pm$1.78
& 24.51$\pm$4.05
& 13.34$\pm$0.72
& 10.14$\pm$0.39
& 12.35$\pm$0.69
& OOM
& 21.83$\pm$3.98
& 8.81 \\

\midrule
\multicolumn{10}{c}{GGAD Methods} \\
\midrule
ARC*~(2024)
& \underline{51.36$\pm$0.14}
& \textbf{29.70$\pm$0.86}
& \textbf{31.44$\pm$1.77}
& \underline{19.79$\pm$0.38}
& 17.31$\pm$0.81
& \underline{19.86$\pm$1.18}
& \underline{32.29$\pm$0.87}
& \underline{40.11$\pm$3.90}
& \underline{2.00} \\

IA-GGAD*~(2025)
& 35.21$\pm$0.33
& 10.57$\pm$1.58
& 13.40$\pm$1.65
& 18.82$\pm$0.71
& \underline{18.59$\pm$0.51}
& 17.67$\pm$1.10
& 32.26$\pm$0.63
& 12.49$\pm$1.57
& 7.50 \\

AnomalyGFM*~(2025)
& 28.61$\pm$0.95
& 13.88$\pm$3.92
& 16.81$\pm$3.21
& 7.46$\pm$0.83
& 6.46$\pm$0.26
& 7.25$\pm$0.42
& 8.34$\pm$0.81
& 29.45$\pm$3.44
& 10.38 \\

UNPrompt*~(2025)
& 12.04$\pm$1.11
& 9.98$\pm$2.40
& 18.23$\pm$4.92
& 6.07$\pm$0.27
& 6.60$\pm$0.24
& 5.81$\pm$0.23
& 13.22$\pm$3.14
& 24.83$\pm$1.67
& 11.88 \\

OWLEYE*~(2026)
& 43.90$\pm$0.01
& 21.98$\pm$0.02
& 27.93$\pm$0.09
& 12.95$\pm$0.40
& 17.72$\pm$0.01
& 17.26$\pm$0.01
& 22.38$\pm$0.01
& 35.23$\pm$0.11
& 3.75 \\

ProMoS~(2026)
& 25.65$\pm$0.94
& 12.21$\pm$0.27
& 22.86$\pm$0.20
& 12.24$\pm$0.34
& 12.59$\pm$0.25
& 12.28$\pm$0.58
& 29.61$\pm$1.62
& 29.83$\pm$1.02
& 6.25 \\

\midrule
\multicolumn{10}{c}{TAGAD Methods} \\
\midrule
CMUCL~(2025)
& 14.29$\pm$0.66
& 5.90$\pm$0.32
& 13.61$\pm$3.48
& 5.49$\pm$0.20
& 5.50$\pm$0.07
& 5.27$\pm$0.04
& 5.17$\pm$0.10
& 15.03$\pm$3.28
& 15.62 \\

CoLL~(2025)
& 7.93$\pm$0.72
& 19.99$\pm$1.70
& 15.01$\pm$1.84
& 11.17$\pm$0.74
& 13.18$\pm$1.47
& 11.12$\pm$1.38
& 16.98$\pm$0.95
& 22.85$\pm$1.97
& 8.50 \\

ProTAGAD
& \textbf{52.43$\pm$0.97}
& \underline{27.20$\pm$0.69}
& \underline{29.65$\pm$0.53}
& \textbf{21.92$\pm$0.12}
& \textbf{19.17$\pm$0.36}
& \textbf{20.30$\pm$1.00}
& \textbf{41.04$\pm$0.43}
& \textbf{44.88$\pm$2.88}
& \textbf{1.25} \\

\bottomrule
\end{tabular}
}
\caption{Performance comparison on different target domains (AUPRC, \%, mean $\pm$ std). ``Rank'' is the average rank over eight targets; Methods marked with * are supervised methods. The best and second-best results are shown in bold and underlined, respectively. OOM denotes out of memory.}
\label{tab:main_results_ap}
\end{table*}
\section{Performance Comparison of AUPRC}

As shown in Table~\ref{tab:main_results_ap}, ProTAGAD achieves the best overall AUPRC performance on the eight unseen target graphs, ranking first on six datasets and second on the remaining two, with the lowest average rank of 1.25. Compared with the strongest baseline ARC, ProTAGAD improves the average AUPRC from 30.23\% to 32.07\%. The largest gains are observed on Fitness (+8.75\%) and Cornell (+4.77\%), while consistent improvements on Citeseer, Grocery, Movies, and Toys demonstrate its effectiveness across citation, e-commerce, and web domains. Although ProTAGAD is slightly inferior to ARC on Pubmed and Texas, the gaps remain limited to 2.50\% and 1.79\%, respectively. Overall, these results indicate that separately preserving textual anomaly cues and topological normality patterns enables ProTAGAD to maintain robust anomaly discrimination under diverse graph structures and anomaly distributions, further validating the effectiveness of the decoupled prototype design in alleviating the \textit{BAB} issue.

\section{Effect of GNN Propagation on $\mathcal{ABS}$}
\begin{figure*}[t]
    \centering

    \begin{minipage}[t]{0.24\textwidth}
        \centering
        \includegraphics[width=\linewidth]{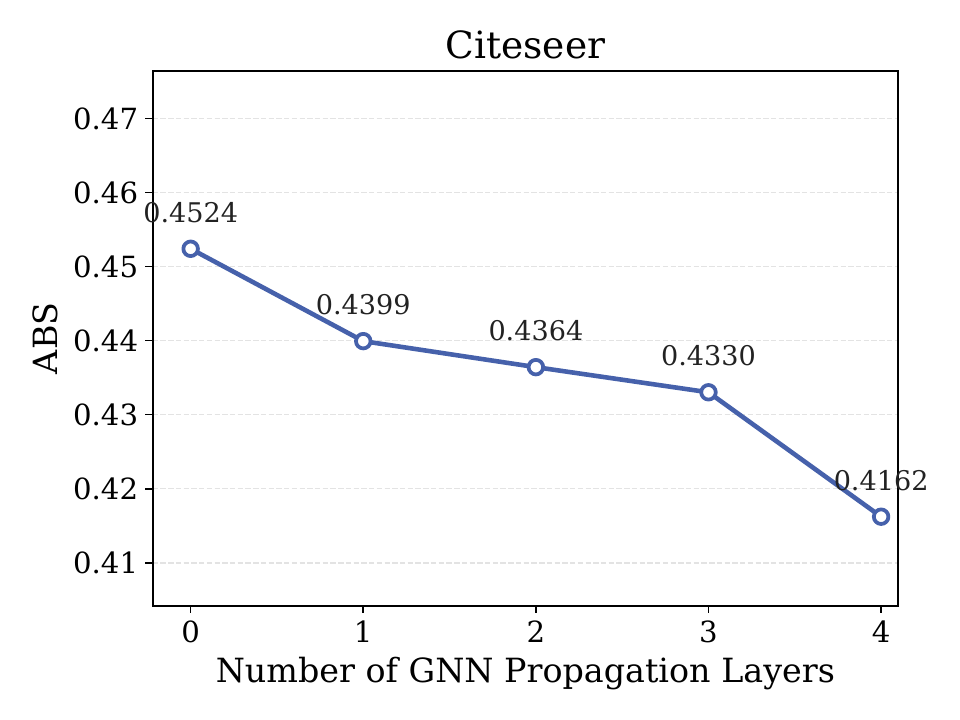}
        \vspace{-2mm}
        \centerline{\small (a) Citeseer}
    \end{minipage}
    \hfill
    \begin{minipage}[t]{0.24\textwidth}
        \centering
        \includegraphics[width=\linewidth]{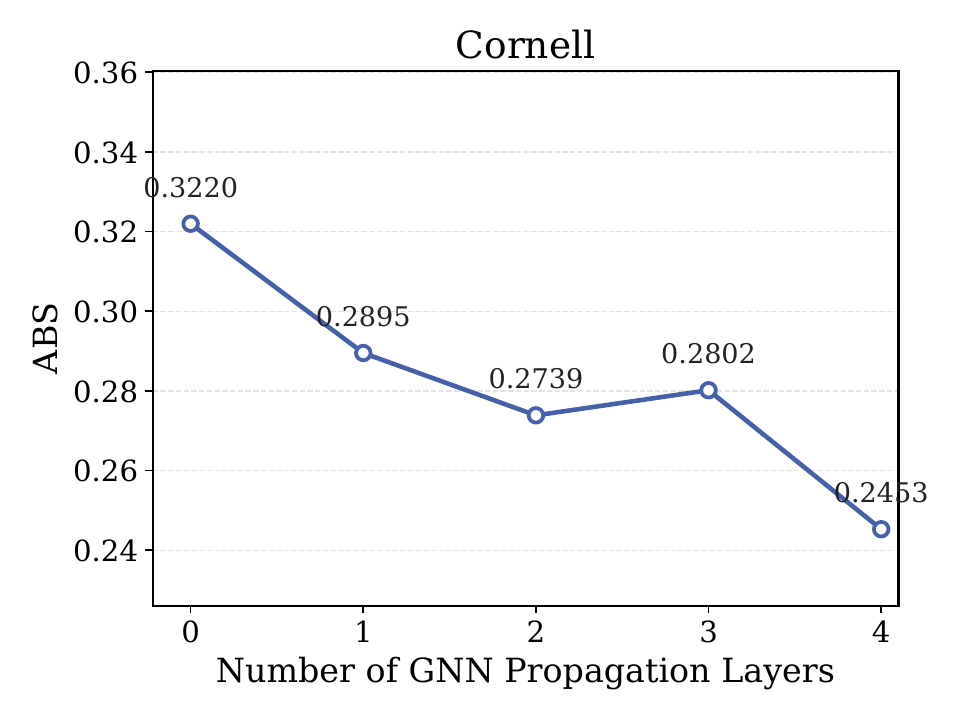}
        \vspace{-2mm}
        \centerline{\small (b) Cornell}
    \end{minipage}
    \hfill
    \begin{minipage}[t]{0.24\textwidth}
        \centering
        \includegraphics[width=\linewidth]{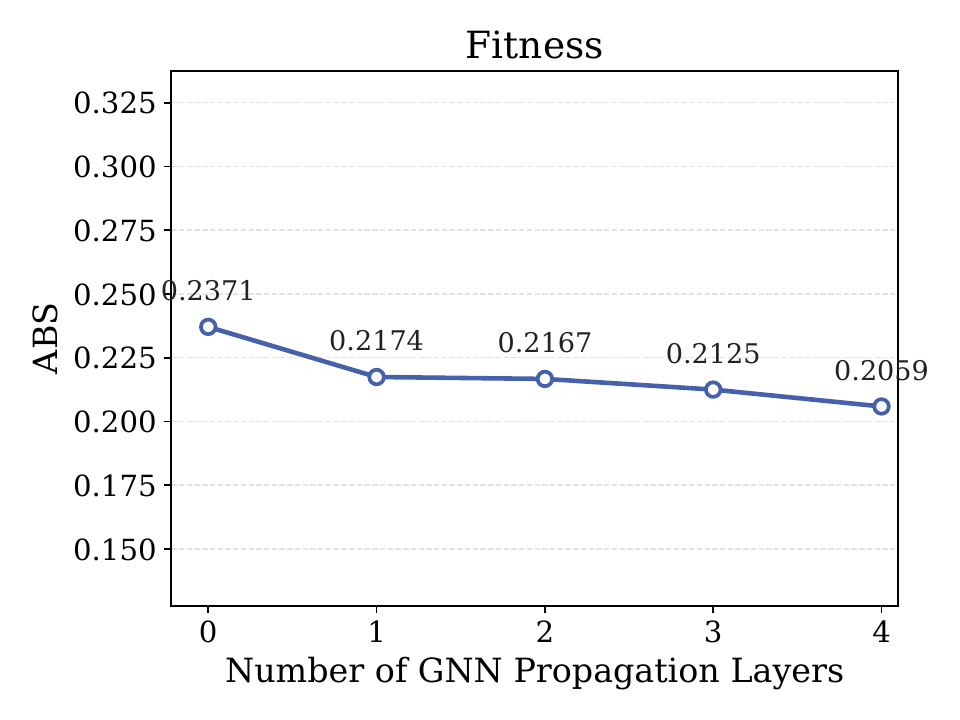}
        \vspace{-2mm}
        \centerline{\small (c) Fitness}
    \end{minipage}
    \hfill
    \begin{minipage}[t]{0.24\textwidth}
        \centering
        \includegraphics[width=\linewidth]{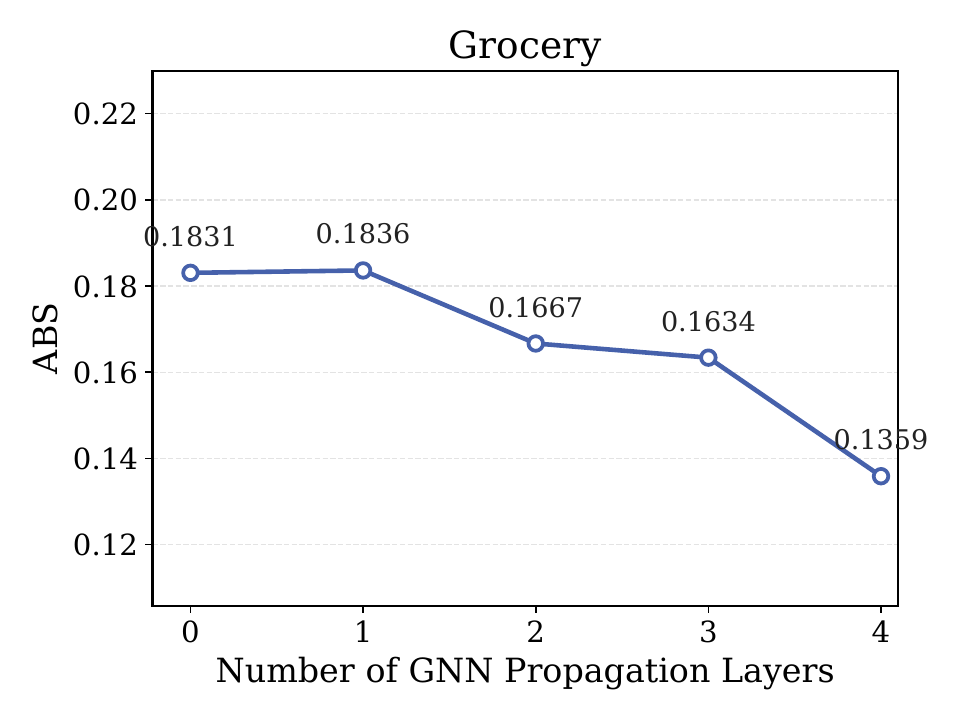}
        \vspace{-2mm}
        \centerline{\small (d) Grocery}
    \end{minipage}

    \vspace{1.5mm}

    \begin{minipage}[t]{0.24\textwidth}
        \centering
        \includegraphics[width=\linewidth]{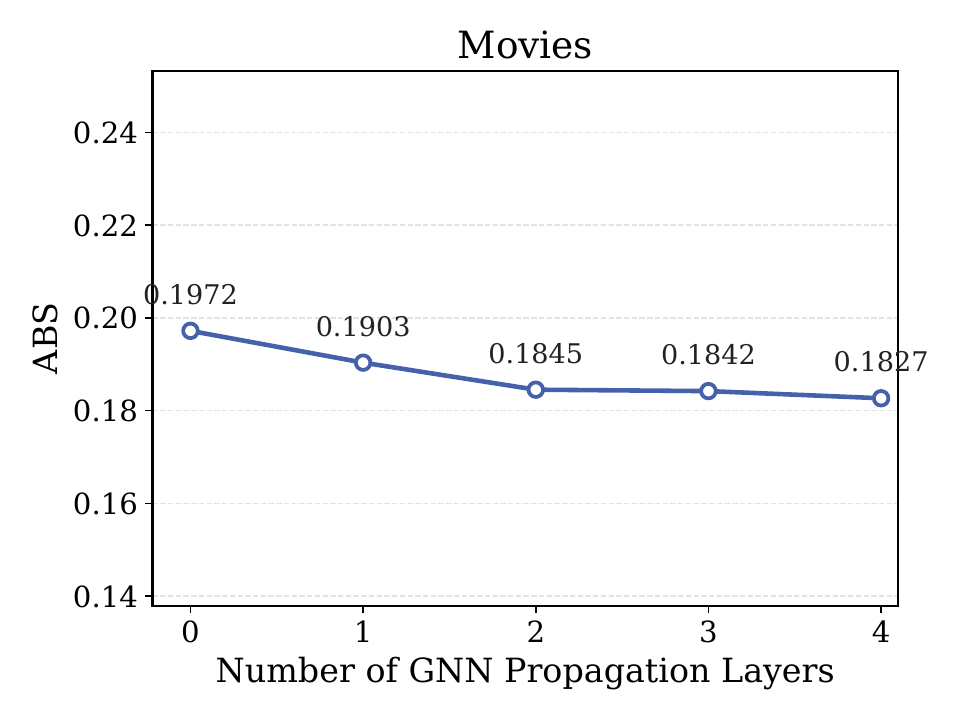}
        \vspace{-2mm}
        \centerline{\small (e) Movies}
    \end{minipage}
    \hfill
    \begin{minipage}[t]{0.24\textwidth}
        \centering
        \includegraphics[width=\linewidth]{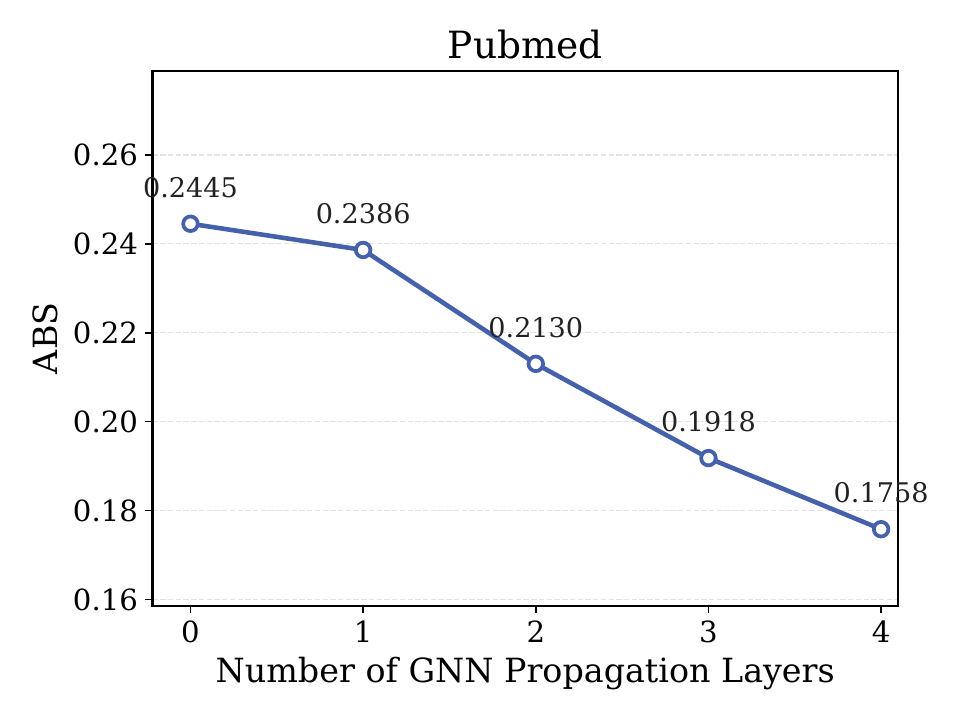}
        \vspace{-2mm}
        \centerline{\small (f) Pubmed}
    \end{minipage}
    \hfill
    \begin{minipage}[t]{0.24\textwidth}
        \centering
        \includegraphics[width=\linewidth]{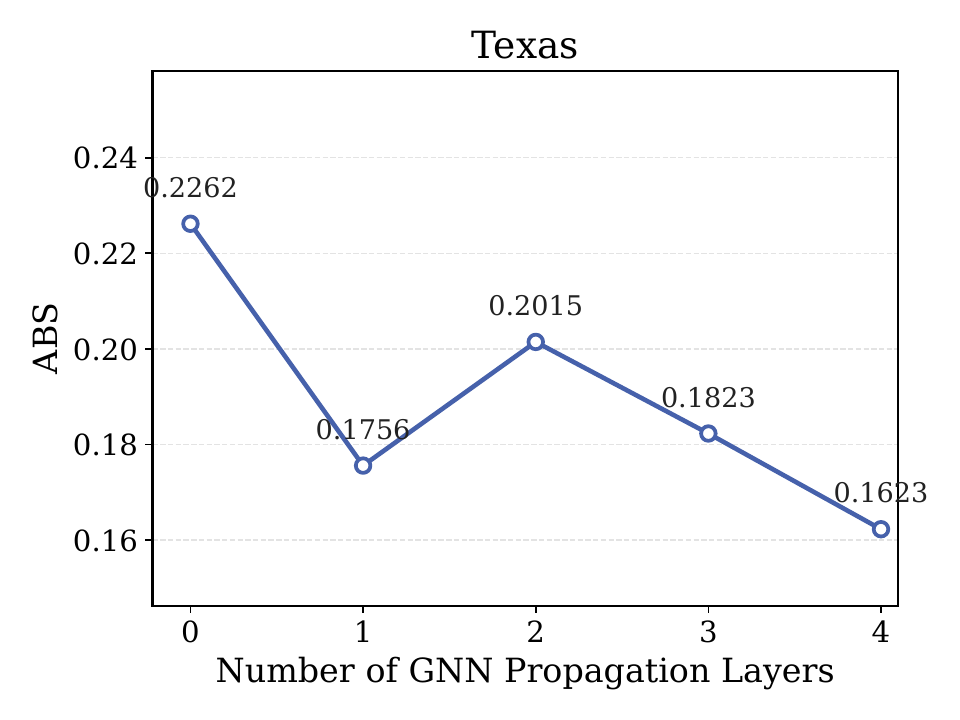}
        \vspace{-2mm}
        \centerline{\small (g) Texas}
    \end{minipage}
    \hfill
    \begin{minipage}[t]{0.24\textwidth}
        \centering
        \includegraphics[width=\linewidth]{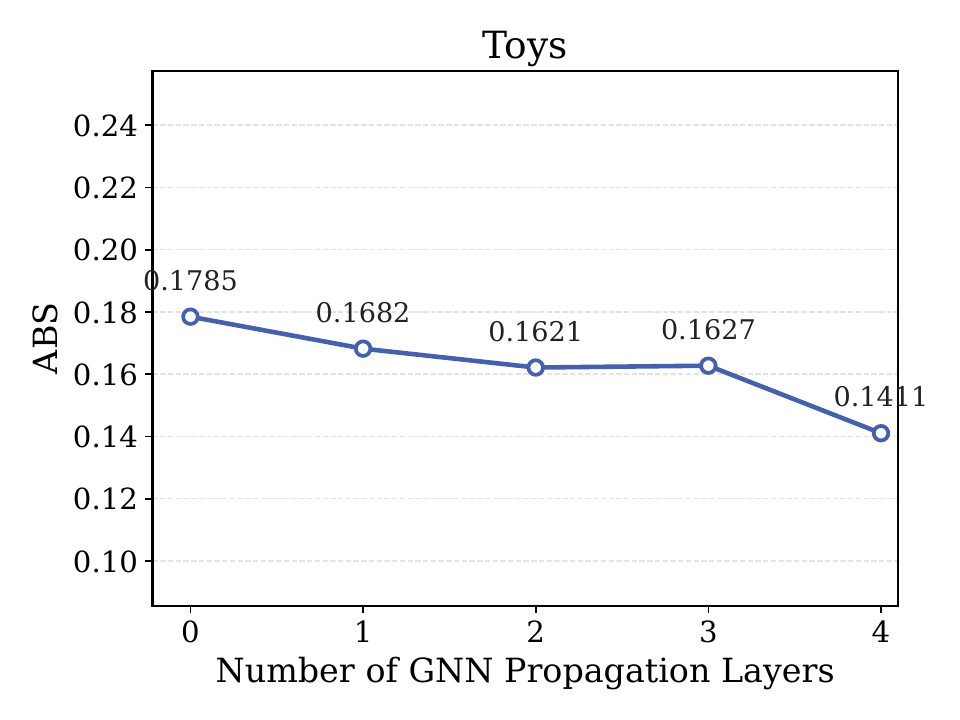}
        \vspace{-2mm}
        \centerline{\small (h) Toys}
    \end{minipage}

    \caption{Effect of coupled GNN propagation on Anomaly Boundary Separability ($\mathcal{ABS}$) across eight target graphs. Layer 0 denotes the anomaly scores before graph propagation, while Layers 1--4 denote the scores after successive propagation layers. A lower $\mathcal{ABS}$ indicates a more severe blurred-anomaly-boundary issue.}
    \label{fig:propagation_abs}
\end{figure*}
To directly examine whether graph propagation induces the blurred-anomaly-boundary issue, we evaluate the coupled variant with
zero to four GNN propagation layers while keeping all other settings unchanged. Layer 0 measures $\mathcal{ABS}$ before graph propagation, whereas Layers 1--4 measure it after successive propagation steps. As shown in Figure~\ref{fig:propagation_abs}, the four-layer
$\mathcal{ABS}$ is lower than its pre-propagation value on all eight target graphs. On average, $\mathcal{ABS}$ decreases from 0.2551 at Layer 0 to 0.2082 at Layer 4, corresponding to a relative reduction of 18.41\%. Particularly pronounced reductions are observed on Texas, Pubmed, and Grocery, where $\mathcal{ABS}$ decreases by 28.25\%, 28.10\%, and 25.78\%, respectively. Although several datasets exhibit minor intermediate fluctuations, the overall tendency remains downward, and every dataset obtains its lowest or near-lowest separability after deeper propagation. Since a lower $\mathcal{ABS}$
indicates greater overlap between the normal and anomalous score distributions, these results provide direct evidence that repeated coupled propagation weakens anomaly boundary separability and aggravates the BAB issue. This observation further motivates the decoupled design of ProTAGAD, which avoids repeatedly propagating textual anomaly cues through graph neighborhoods and preserves their discriminability before decision-level fusion.

\section{Algorithmic description}
The algorithmic description of the training and inference process of ProTAGAD is summarized in Algorithm~\ref{alg:training}, and Algorithm~\ref{alg:inference}, respectively.
\begin{algorithm}[t]
\renewcommand{\algorithmicrequire}{\textbf{Input:}}
\renewcommand{\algorithmicensure}{\textbf{Parameters:}}
\caption{Training Algorithm of ProTAGAD}
\label{alg:training}
\begin{algorithmic}[1]
\REQUIRE Source graphs $\mathcal{T}_{\mathrm{train}}$.
\ENSURE Training epochs $T$; number of topological prototypes $K$.
\STATE Initialize all learnable parameters.
\FOR{each epoch $t=1,\ldots,T$}
    \FOR{each graph $\mathcal{G}\in\mathcal{T}_{\mathrm{train}}$}
        \STATE Derive textual representations $\mathbf{X}^{t}$ via Eq.~(2).
        \STATE Estimate anomaly probabilities $\widehat{\mathbf{Y}}$ via Eq.~(3).
        \STATE Obtain agent-derived pseudo-labels $\mathbf{C}$ via Eq.~(4).
        \STATE Compute the probability estimation loss $\mathcal{L}_{\mathrm{prob}}$ via Eq.~(5).
        \STATE Construct textual prototypes $\mathbf{P}^{t}$ via Eq.~(6).
        \STATE Compute the textual prototype alignment loss $\mathcal{L}_{\mathrm{align}}$ via Eq.~(7).
        \STATE Derive topology-aware representations $\mathbf{H}$ via Eq.~(10).
        \STATE Derive projected representations $\mathbf{H}'$ via Eq.~(11).
        \STATE Construct topological prototypes $\mathbf{P}^{s}$ from $\mathbf{H}$ via Eq.~(12).
        \STATE Compute the structural consistency loss $\mathcal{L}_{\mathrm{str}}$ via Eq.~(13).
        \STATE Update model parameters using $\mathcal{L}_{\mathrm{prob}}$, $\mathcal{L}_{\mathrm{align}}$, and $\mathcal{L}_{\mathrm{str}}$.
    \ENDFOR
\ENDFOR
\end{algorithmic}
\end{algorithm}

\begin{algorithm}[t]
\renewcommand{\algorithmicrequire}{\textbf{Input:}}
\renewcommand{\algorithmicensure}{\textbf{Parameters:}}
\caption{Zero-Shot Inference Algorithm of ProTAGAD}
\label{alg:inference}
\begin{algorithmic}[1]
\REQUIRE Target graphs $\mathcal{T}_{\mathrm{test}}$; trained model; textual prototypes $\mathbf{P}^{t}$; topological prototypes $\mathbf{P}^{s}$.
\ENSURE Well-trained ProTAGAD model weight parameters.
\FOR{each graph $\mathcal{G}\in\mathcal{T}_{\mathrm{test}}$}
    \STATE Derive textual representations $\mathbf{X}^{t}$ and anomaly probabilities $\widehat{\mathbf{Y}}$ via Eqs.~(2) and~(3).
    \STATE Compute the textual anomaly score $S^{t}(v_i)$ via Eq.~(9).
    \STATE Derive node representations $\mathbf{H}$ and $\mathbf{H}'$ via Eqs.~(10) and~(11).
    \STATE Compute the topological anomaly score $S^{s}(v_i)$ via Eq.~(14).
    \STATE Compute the final anomaly score $S(v_i)$ via Eq.~(15).
\ENDFOR
\STATE \textbf{return} $\{S(v_i)\}$.
\end{algorithmic}
\end{algorithm}


\end{document}